\documentclass[journal]{IEEEtran}

\makeatletter
\def\endthebibliography{%
  \def\@noitemerr{\@latex@warning{Empty `thebibliography' environment}}%
  \endlist
}
\makeatother

\usepackage{lineno}
\usepackage{color}
\usepackage{graphicx}
\usepackage{cite}
\usepackage{amsfonts}
\usepackage[cmex10]{amsmath}
\usepackage{amsmath}
\usepackage{caption}
\usepackage{amssymb}

\usepackage{booktabs}
\usepackage{multirow}
\usepackage{times}
\usepackage{soul}
\usepackage{indentfirst}
\usepackage{booktabs}

\newcommand{\etal}{\textit{et al.}}

\usepackage[pagebackref=true,breaklinks=true,letterpaper=true,colorlinks,bookmarks=false,citecolor=cyan]{hyperref}

\ifCLASSINFOpdf

\else

\fi

\begin{document}

\title{Unsupervised High-Resolution Portrait Gaze Correction and Animation}

\author{Jichao~Zhang,
        Jingjing~Chen,
        Hao~Tang,
        Enver~Sangineto, \\ 
        Peng~Wu,
        Yan Yan,
        Nicu~Sebe, \textit{Senior Member, IEEE},
        Wei~Wang
        \IEEEcompsocitemizethanks{

\IEEEcompsocthanksitem Jichao Zhang, Nicu Sebe, and Wei Wang are with the Department of Information Engineering and Computer Science (DISI), University of Trento, Italy. E-mail: jichao.zhang@unitn.it.
\IEEEcompsocthanksitem Jingjing Chen and Peng Wu are from the Zhejiang University and the Shandong University of China, respectively.
\IEEEcompsocthanksitem Hao Tang is with the Computer Vision Lab, ETH Zurich, Switzerland. 
E-mail: hao.tang@vision.ee.ethz.ch. 
\IEEEcompsocthanksitem Enver~Sangineto is with the Department of Engineering (DIEF), University of Modena and Reggio Emilia, Modena, Italy. E-mail: enver.sangineto@unimore.it.
\IEEEcompsocthanksitem Yan Yan is with the Department of Computer Science of Illinois Institute of Technology, United States. E-mail: yyan34@iit.edu
}
}

\markboth{IEEE TRANSACTION ON IMAGE PROCESSING}%
{Shell \MakeLowercase{\textit{et al.}}: Bare Demo of IEEEtran.cls for Journals}

\maketitle


% \input{sections/1-abstract}

\begin{abstract}
This paper proposes a gaze correction and animation method for high-resolution, unconstrained portrait images, which can be trained without the gaze angle and the head pose annotations. Common gaze-correction methods usually require annotating training data with precise gaze, and head pose information. Solving this problem using an unsupervised method remains an open problem, especially for high-resolution face images in the wild, which are not easy to annotate with gaze and head pose labels. To address this issue, we first create two new portrait datasets: CelebGaze ($256 \times 256$) and high-resolution CelebHQGaze ($512 \times 512$). Second, we formulate the gaze correction task as an image inpainting problem, addressed using a Gaze Correction Module~(GCM) and a Gaze Animation Module~(GAM). Moreover, we propose an unsupervised training strategy, i.e., Synthesis-As-Training, to learn the correlation between the eye region features and the gaze angle. As a result, we can use the learned latent space for gaze animation with semantic interpolation in this space. Moreover, to alleviate both the memory and the computational costs in the training and the inference stage, we propose a Coarse-to-Fine Module~(CFM) integrated with GCM and GAM. Extensive experiments validate the effectiveness of our method for both the gaze correction and the gaze animation tasks in both low and high-resolution face datasets in the wild and demonstrate the superiority of our method with respect to the state of the arts.
\end{abstract}


\IEEEpeerreviewmaketitle

\section{Introduction}

The goal of the gaze correction task is to manipulate the gaze direction of a face image with respect to a specific target direction. The main application of this task is altering the eye appearance so that the person's gaze is directed into the camera. For example, shooting a good portrait is challenging as the subjects may be too nervous to stare at the camera. Another scenario is videoconferencing, where eye contact is very important. The gaze can express attributes such as attentiveness and confidence. Unfortunately, eye contact is frequently lost during a video conference, as the participants look at the monitors and not directly into the camera. Moreover, some works use gaze redirection to improve few-shot gaze estimation task~\cite{yu2019improving,yu2020unsupervised}.

\begin{figure}[t]
\begin{center}
\includegraphics[width=1.0\linewidth]{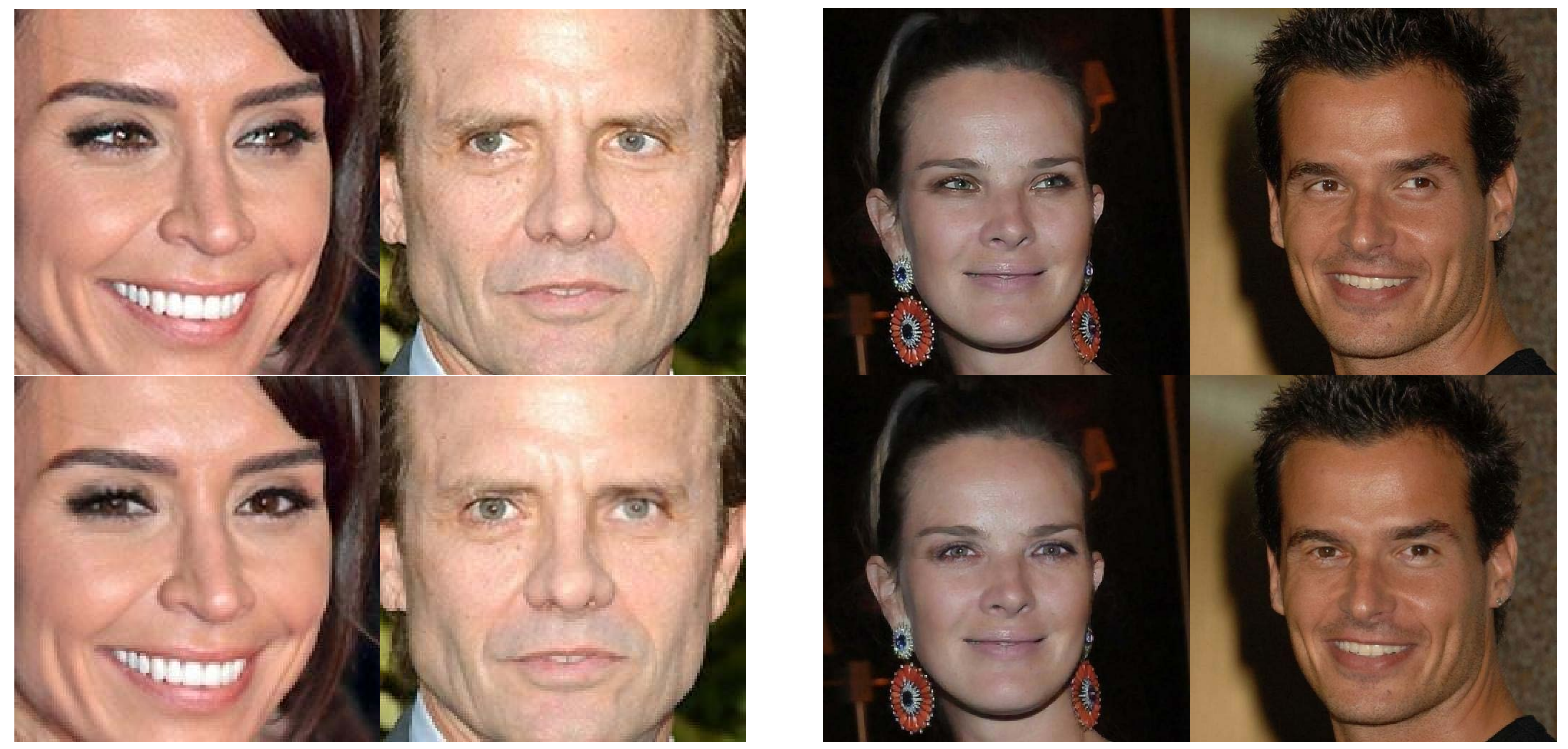}
\end{center}
\vspace{-0.3cm}
\caption{Left: $256 \times 256$ images and the corresponding gaze-corrected results generated by our method using samples of the  CelebGaze dataset. Right: $512 \times 512$ high-resolution images and the gaze-corrected results using samples of our dataset CelebHQGaze. The first and second rows show the original images and eye-gaze corrected results, respectively.}
\label{fig:newgaze}
\vspace{-0.5cm}
\end{figure}

Early works in gaze correction relied on special hardware, such as stereo cameras~\cite{criminisi2003gaze,yang2002eye}, Kinect sensors~\cite{kuster2012gaze} or transparent mirrors~\cite{kollarits199634,okada1994multiparty}. Recently, a few methods based on machine learning showed a good quality synthetic for gaze correction. For instance, Kononenko and Lempitsky~\cite{kononenko2015learning} propose to solve the problem of monocular gaze correction using decision forests. DeepWarp~\cite{ganin2016deepwarp} uses a deep network to directly predict an image-warping flow field with a coarse-to-fine learning process. However, this method fails in generating photo-realistic images when the gaze redirection involves large angles. Moreover, it produces unnatural eye shapes because of the $L1$ loss, which is used to learn the flow field without any geometric-based regularization. To solve this problem, PRGAN~\cite{he2019photo} proposes to exploit adversarial learning with a cycle-consistent loss to generate more plausible gaze redirection results. However, these methods~\cite{kononenko2015learning,ganin2016deepwarp,he2019photo} fail in obtaining high-quality gaze redirection results in the wild when there are large variations in the head pose and the gaze angles. Recently, Marcel \etal~\cite{Buehler2019ICCVW} proposed a content-consistent model for realistic eye generation. However, their approach is based on semantic segmentation masks, which implies a great annotation effort. Another category of works is based on  3D models without training data, such as GazeDirector~\cite{wood2018gazedirector}. The main idea of GazeDirector is to model the eye region in 3D instead of predicting a flow field directly from an input image. However, modeling in 3D has strong assumptions that do not hold in non-laboratory scenarios.

The unsupervised method can avoid expensive annotations. Moreover, it has essential significance for image representation and semantic disentanglement. Thus, we recently proposed a novel gaze-correction method, GazeGAN~\cite{10.1145/3394171.3413981}, which is extended in this paper. In \cite{10.1145/3394171.3413981}, we collected the CelebGaze dataset, which consists of two image domains: $X$, with eyes staring at the camera,  and $Y$, with eyes looking somewhere else (see Fig.~\ref{fig:newgaze}, \textit{left}). Note that the  CelebGaze images do not annotate the gaze angle or the head pose. Moreover, in \cite{10.1145/3394171.3413981} we propose an unsupervised learning method for gaze correction and animation, which consists of two main modules: the Gaze Correction Module (GCM) and the Gaze Animation Module (GAM). GCM is an inpainting model, trained on a domain $X$, which learns how to fill in the missing eye regions with a new content representing the gaze-corrected eyes. GAM is another inpainting model used for gaze animation, and it is trained on a domain $Y$. To generalize the gaze redirection to various angle directions (i.e., in ``animations''), we propose a training method (Synthesis-As-Training) that uses synthetic data to train GAM and encourages the encoded features of the eye region to be correlated with the gaze angle. Then, gaze animation can be achieved by interpolating these features in the latent space. 

\begin{figure}[t]
\begin{center}
\includegraphics[width=1.0\linewidth]{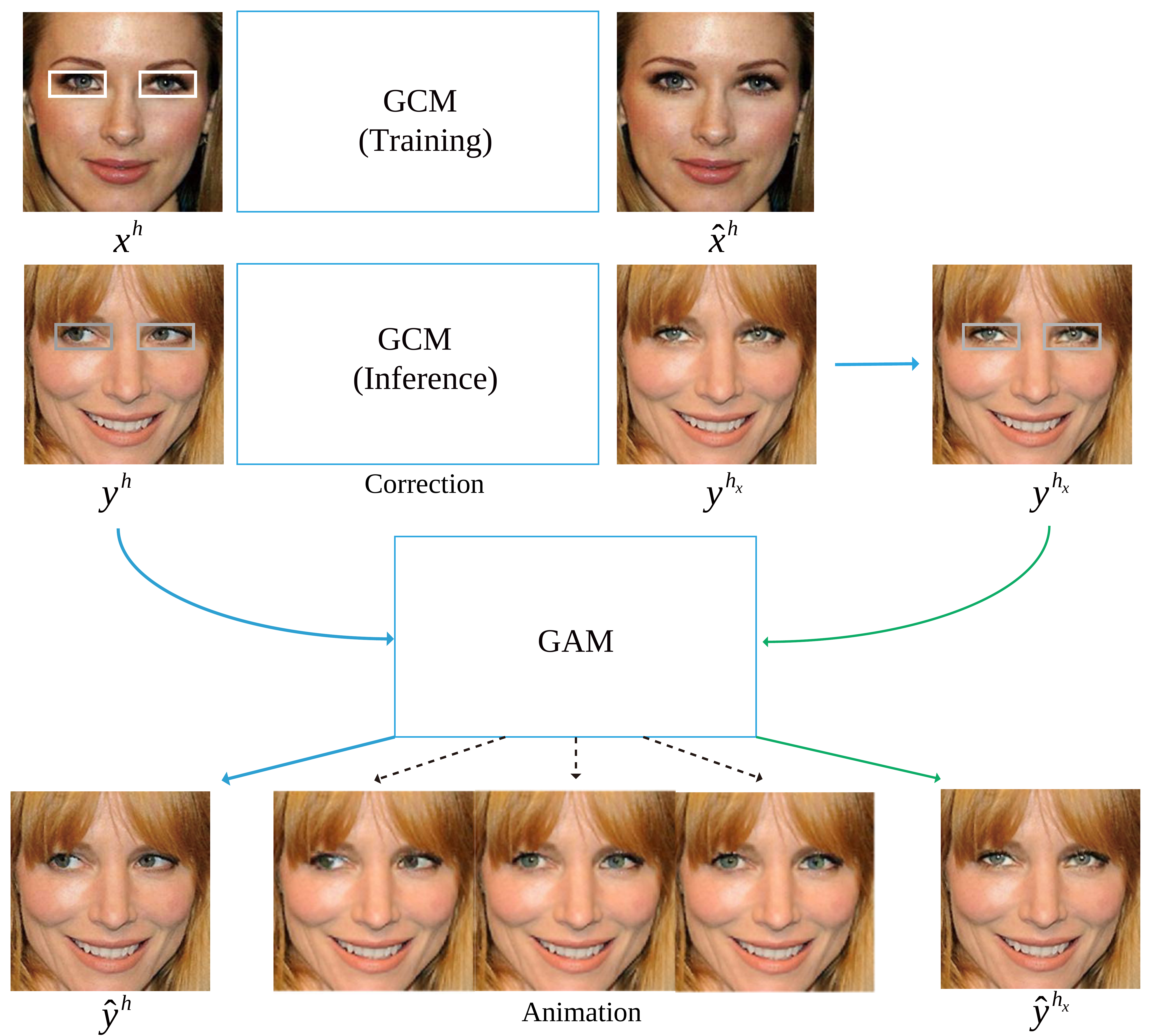}
\end{center}
\vspace{-0.2cm}
\caption{Overview of the proposed architecture. We have two main modules: Gaze Correction Module for performing gaze correction (GCM) and Gaze Animation Module for performing gaze animation (GAM). Moreover, we propose to use the gaze-corrected samples from GCM to train GAM (Synthesis-as-Training). The trained GAM can achieve gaze animation by interpolating the latent feature. The white boxes are the eye mask to remove the eye region. The gray boxes represent the cropping of eye region}
\vspace{-0.8cm}
\label{fig:overview}
\end{figure}

In this paper, we extend  GazeGAN~\cite{10.1145/3394171.3413981} to work also with higher resolution portrait images. Specifically, we first create a new dataset, CelebHQGaze, containing $512 \times 512$ high-resolution portrait images, as shown in Fig.~\ref{fig:newgaze} (\textit{right}). Second, we propose a novel GCM and GAM integrated with a coarse-to-fine module (CFM). In more detail, CFM first allows the inpainting model to be trained using low-resolution images for coarse-grained image generation. Then it uses a global nonparametric model, Laplacian Reconstruction, and a local parametric model, Local-Refinement Autoencoder, to compensate for the high-frequency information loss and to remove possible artifacts for the eye region. Utilizing this new architecture,  we can avoid training each module using high-resolution images. CFM speeds up both the training and the inference process while obtaining high-quality results, comparable with directly training with high-resolution images.

Similar to GazeGAN~\cite{10.1145/3394171.3413981}, an autoencoder is pretrained using self-supervised mirror learning~(PAM), where the bottleneck features are used as an extra input to the dual inpainting model to preserve the identity of the corrected results. Moreover, global and local discriminators are used to improve the visual quality of the generated samples. Finally, our qualitative and quantitative evaluations show that our method generates higher-quality results with respect to the state-of-the-arts in both the gaze correction and the gaze animation tasks.  
\begin{figure*}[t]
\begin{center}
\includegraphics[width=1.0\linewidth]{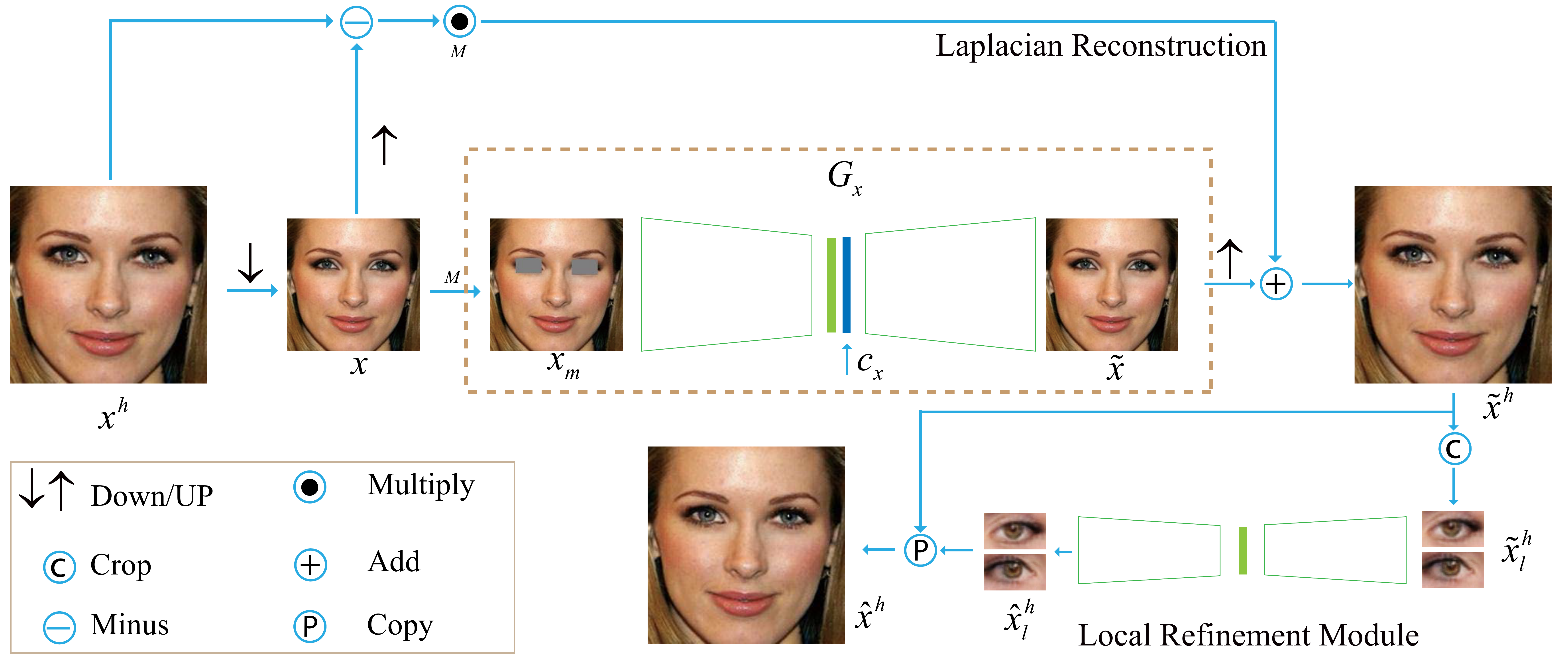}
\end{center}
\vspace{-0.2cm}
\caption{Overview of the architecture for Gaze Correction Module (GCM) integrated with Coarse-to-Fine Module (CFM) which consists of one laplacian reconstruction and one local-refinement module. CFM first allows the inpainting network $G_{x}$ trained using low-resolution images to attain coarse-grained inpainted results, then attains high-resolution results by the global nonparametric Laplacian reconstruction, and finally exploits a parametric local-refinement module (LRM) to compensate for high-frequency information and remove artifacts for the eye region. We use 2$\times$ scales for downsampling and upsampling.}
\vspace{-0.2cm}
\label{fig:gcm}
\end{figure*}

We summarize below our main contributions:
\begin{itemize}
    \item[1)] We introduce an unsupervised inpainting architecture for high-resolution gaze correction and animation. 
    \item[2)] We propose a novel CFM module that can alleviate both the memory and the computational costs in the training and the inference stage while achieving high-quality results comparable with training with high-resolution facial images.
    \item[3)] We propose a gaze animation module and a Synthesis-As-Training method to generate gaze-correction results with variable angles. 
    \item[4)] We make publicly available the  CelebHQGaze dataset for the research community interested in gaze correction and animation: \url{https://github.com/zhangqianhui/GazeAnimationV2}. 
\end{itemize}

\section{Related Work}
{\bfseries Generative Adversarial Networks.}
Generative Adversarial Networks (GANs)~\cite{goodfellow2014generative} are powerful generative models which learn a  distribution that mimics a given target distribution. They have been applied to many fields, such as low-level image processing tasks~(e.g., image inpainting~\cite{pathak2016context,IizukaSIGGRAPH2017}, image super-resolution~\cite{Ledig_2017_CVPR,DRIT,wang2018esrgan}),  semantic and style transfer~(e.g., image translation~\cite{isola2017image,tang2019local,Zhu_2017_ICCV,liu2017face,tang2019multichannel,park2020contrastive,mallya2020world}, image attribute manipulation~\cite{Zhang:2018:SGM:3240508.3240594,he2019attgan,he2020pa,liu2019stgan,chencoogan2020,chusscgan2020}, person image synthesis~\cite{tang2020xinggan,tang2019cycle,PAMIStephane,zhang20213d,zhang2021controllable}, image manipulation~\cite{park2020swapping}). 

{\bfseries Image Inpainting.}
Image Inpainting is an important task in computer vision and computer graphics, and it aims to
fill in the missed/masked pixels of an image utilizing plausible synthesized content. Most of the previous methods can be split into two classes. The first is based on diffusion or patch-based approaches, which rely on handcrafted low-level features. For example,
PatchMatch~\cite{barnes2009patchmatch} is a fast nearest-neighbor field algorithm, which can perform real-time  image inpainting. Generally speaking, this class of methods is based on low-level features. They are usually ineffective in filling in the missing part of an image when the underlying semantic
structure is not trivial and cannot generate novel objects that cannot be found in other non-masked parts of the source image. The second class of methods is based on learning approaches. Recently, CNN-based and GAN-based methods have shown promising performance on image inpainting~\cite{pathakCVPR16context,iizuka2017globally,liu2019coherent,zhang2018semantic}. For instance, inpainting can be used for facial attributes manipulation such as hair, mouth, and eyes~\cite{Jo_2019_ICCV,dolhansky2018eye,olszewski2020intuitive}. 
We also adopt an inpainting approach, differently from previous work,
our method does not require the data to be labeled with additional information, such as semantic labels, sketches, or reference images. 

{\bfseries Gaze Correction.}
Previous work for gaze correction can be split into three main classes: 1) hardware-driven, 2) rendering and synthesis, 3) learning-based. 

The hardware support was indispensable in early research. Kollarits~\etal~\cite{kollarits199634}  use half-silvered mirrors to place
the camera on the optical path of the display. Yang~\etal~\cite{yang2002eye} address the eye contact problem with a 
view synthesis, and they use a pair of calibrated stereo cameras jointly with a face model to track the head pose in 3D. Generally speaking, these hardware-based methods are expensive.

The second class of approaches typically renders the eye region based on a 3D fitting model, which replaces the original eyes with synthetic eyeballs. Banf~\etal~\cite{banf2009example} use an example-based approach for deforming the eyelids and sliding the iris across
the model surface with a texture-coordinate interpolation. To fix the limitations caused by the use of a mesh, where the face and eyes are mixed, GazeDirector~\cite{wood2018gazedirector}
separately deals with the face and eyeballs, synthesizing more high-quality images, especially for large redirection angles. These methods usually struggle in realistically rendering the corrected eyes. Additionally, modeling methods have strong assumptions that usually do not hold in practice.

Concerning the third class of methods, the core idea for most of the learning-based approaches is to use a large paired training dataset to train a statistical model~\cite{kononenko2015learning,kononenko2017photorealistic,yu2019improving,park2019few,chen2021coarse}. Some methods~\cite{kononenko2015learning,kononenko2017photorealistic} learn to generate the flow field, which is then used to relocate the eye pixels in the original image. For instance, Ganin~\etal~\cite{ganin2016deepwarp}
use a CNN to learn the flow field, which warps the input image and redirects the gaze to the target angle. However, \cite{ganin2016deepwarp} fails to generate photo-realistic and natural shapes because it uses only pixel-wise differences between the input and the ground truth as the training loss. To address this problem, He \etal~\cite{he2019photo}  use adversarial learning, jointly with a cycle-consistent loss, which can improve the visual quality and the redirection precision. However, these methods can hardly generate plausible results in the wild, i.e., in a scenario with large variations in the head pose, the gaze angle, or the illumination conditions. In contrast, we propose to use dual inpainting modules (GCM and GAM) to correct the gaze angle and achieve high-resolution and high-quality gaze redirection in the wild. 

\begin{figure*}[t]
\begin{center}
\includegraphics[width=1.0\linewidth]{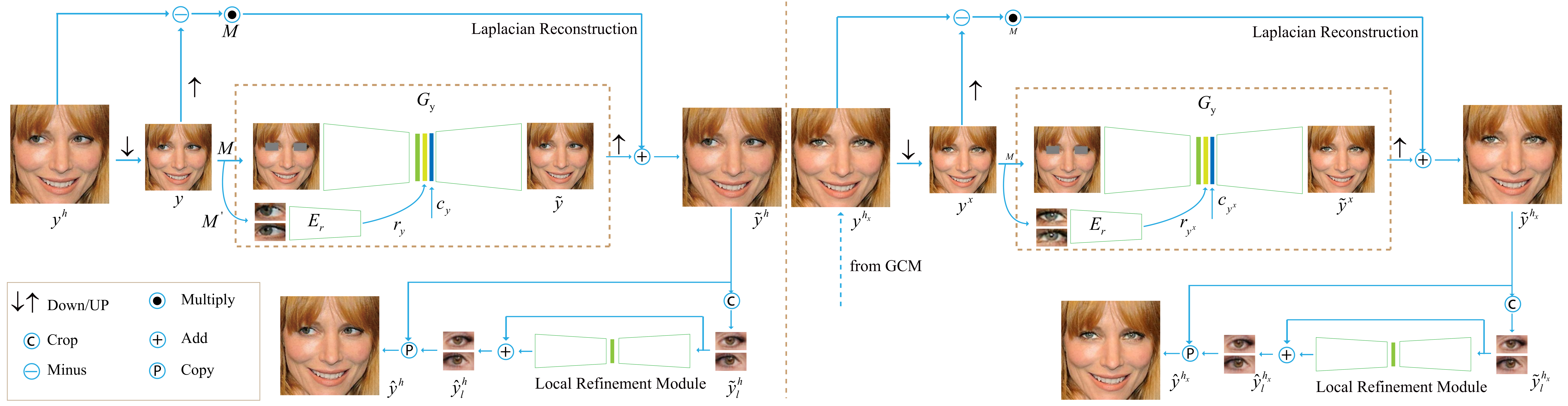}
\end{center}
\vspace{-0.2cm}
\caption{An overview of the proposed Gaze Animation Module (GAM) integrated with Coarse-to-Fine Module (CFM). In the left, $G_{y}$ uses the sample $y \in Y$ for training. Compared to $G_{x}$, the decoder of $G_{y}$ has an extra input $r$ which is provided by the encoder $E_{r}$. In the right, we use GCM to generate the gaze-corrected image $y^{h_x}$, which is then used for training $G_{y}$~(Synthesis-as-Training). With the paired samples $y^{h}$ and $y^{h_{x}}$ for training $G_{y}$, the feature from $E_{r}$ would be correlated with gaze angle, and gaze animation can be achieved by interpolating the feature.} 
\label{fig:gam}
\vspace{-0.4cm}
\end{figure*}

\section{Method}
The overview of our method is shown in Fig.~\ref{fig:overview} and our model consists of two main modules: Gaze Correction Module and Gaze Animation Module. Specifically, Fig. \ref{fig:gcm} illustrates Gaze Correction Module (GCM), integrating with Coarse-to-Fine Module (CFM), which is trained using the sample $x$ from domain $X$. Fig. \ref{fig:gam} illustrates Gaze Animation module (GAM), integrating with CFM, which are trained using the sample $y$ from domain $Y$, and GAM exploits the corrected samples for training to make the eye feature correlate with the gaze angle (Synthesis-as-Training method). Additionally, Fig.~\ref{fig:pre} shows the pretrained autoencoder (PAM), which extracts the angle-invariant content feature as the additional input of GCM and GAM. We 
here clarify the adopted notations.

$\bullet$ $x \in R^{m \times n \times 3}$ is  an image instance, where $m$ and $n$ are the image height and  width, and 3 is the number of RGB channels. 

$\bullet$ The training set is split into two domains: $X$,  containing images with a gaze staring at the camera, and $Y$, containing images with a gaze staring somewhere else. $X^{h}$ and $Y^{h}$ correspond to the higher-resolution sets of $X$ and $Y$, respectively. 

$\bullet$ $M \in R^{m \times n \times 3}$ denotes a binary mask function of the eye region and $M^{'}$ defines the operation of extracting a rectangular sub-image~(the eye region). 

$\bullet$ $P_{x}$ and  $P_{y}$ denote the data distributions in $X$ and $Y$, respectively.  $P_{m}$ indicates the  distribution of the masked data $M(x)$, where the eye region is removed from $x$. If $x \in X$ and $y \in Y$, both $M(x)$ and $M(y)$ have the same distribution, because the only difference between $x$ and $y$ is in the eye region. Thus, $M(x) \thicksim P_{m}$ and $M(y) \thicksim P_{m}$. 

$\bullet$ $r \in R^{128}$ and $c \in R^{256}$ denote the angle, the content features~(being the latter angle invariant), respectively and different encoders extract them.

$\bullet$ $F$ denotes the image horizontal flipping operation (mirroring). 

We first introduce the details of our coarse-to-fine module (CFM).

\subsection{Coarse-to-Fine Module}

In order to alleviate the memory costs and reduce the number of overall training parameters while simultaneously being able to generate high-resolution facial images, we propose a CFM for GCM and GAM. This module consists of a global nonparametric Laplacian Reconstruction for the inpainting process and a local parametric Local Refinement Module (LRM) which will be introduced with details, taking GCM as an example.

\subsubsection{Global Nonparametric Laplacian Reconstruction}

As shown in Fig.~\ref{fig:gcm}, the high-resolution image $x^{h}$ is downsampled by a factor of 2, obtaining $x$, where the latter is as input to the inpainting networks of GCM. The generated image is $\tilde x$. Let $u(.)$ be an upsampling operator which smooths and expands $x$ to the original size (i.e., the resolution of $x^{h}$). The single-level Laplacian pyramid $p$ can be obtained by: 
\begin{equation} \label{pyramid}
\begin{aligned}
p = x^{h} - u(x).
\end{aligned}
\end{equation}

Then, the reconstruction process for the high-resolution image $\tilde x^{h}$ is:
\begin{equation} \label{pyramid}
\begin{aligned}
\tilde x^{h} = u(\tilde x) + M(p),
\end{aligned}
\end{equation}
where we use $M$ to remove the eye region from $p$ which is replaced by the zero. Then we introduce the local refinement process to improve the visual quality and remove the artifacts of the eye regions. 

\subsubsection{Local Parametric LRM}
We use $M^{'}(\tilde x^{h})$ to extract the
local eye region $\tilde x^{h}_{l}$. Then, we utilize one autoencoder $G_{h}$ together with residual image learning, to refine $\tilde x^{h}_{l}$ and get $\hat x^{h}_{l}$. 
Finally, the high-resolution complete image $\hat x^{h}$ can be obtained by replacing the local eye region $\tilde x^{h}_{l}$ with $\hat x^{h}_{l}$.

\subsection{Gaze Correction Module}
\label{GCM}

As shown in Fig.~\ref{fig:gcm}, we first downsample $x_{h}$ to attain the low-resolution $x$, and then take $x$ as the input of inpainting network $G_{x}$ whose goal is to fill in the masked eye region of $x_{m} = M(x)$ by generating the missing eyes. This can be formulated as:
\begin{equation}
\begin{aligned}
& c_{x} = E_{c}(M^{'}(x)), \tilde x = G_{x}(M(x), c_{x}),
\end{aligned}
\end{equation}
where $c_{x}$ are the content (angle-invariant) features encoded using only the eye regions as input ($M^{'}(x)$) of the content encoder $E_{c}$. $E_{c}$ is the encoder of $G_{pre}$ which will be introduced in Sec.~\ref{PAM}.

In principle, $G_{x}$ can learn the mapping from $M(x) \thicksim P_{m}$ to $x \thicksim P_{x}$ by training. Given one sample $y \thicksim P_{y}$, then, we remove the eye region to get $M(y) \thicksim P_{m}$, because $x$ and $y$ have different distributions only in the eye region. Thus $G_{x}$ can be used to map $M(y)$ into the $G_{x}(M(y)) \thicksim P_{x}$ which is the intuitive basis of our correction module. This can be formulated as:
\begin{equation}
\begin{aligned}
c_{y} = E_{c}(M^{'}(y)), y^{x} = G_{x}(M(y), c_{y}),
\end{aligned}
\end{equation}
where $c_{y}$ are the content (angle-invariant) features encoded using only the eye regions as input $M^{'}(y)$ of the content encoder $E_{c}$. 

We train the GCM using high-resolution face images based on the CFM module. At the inference time, for the corrected result  $y^{x}$, we get a high-resolution result $y^{h_{x}}$ by compensating for the high-frequency details using the laplacian reconstruction and LRM. Note that the corrected result $y^{h_{x}}$ is also used for training GAM. More details can be found below.

\subsection{Gaze Animation Module}
\label{GAM}

\begin{figure}[t]
\begin{center}
\includegraphics[width=1.0\linewidth]{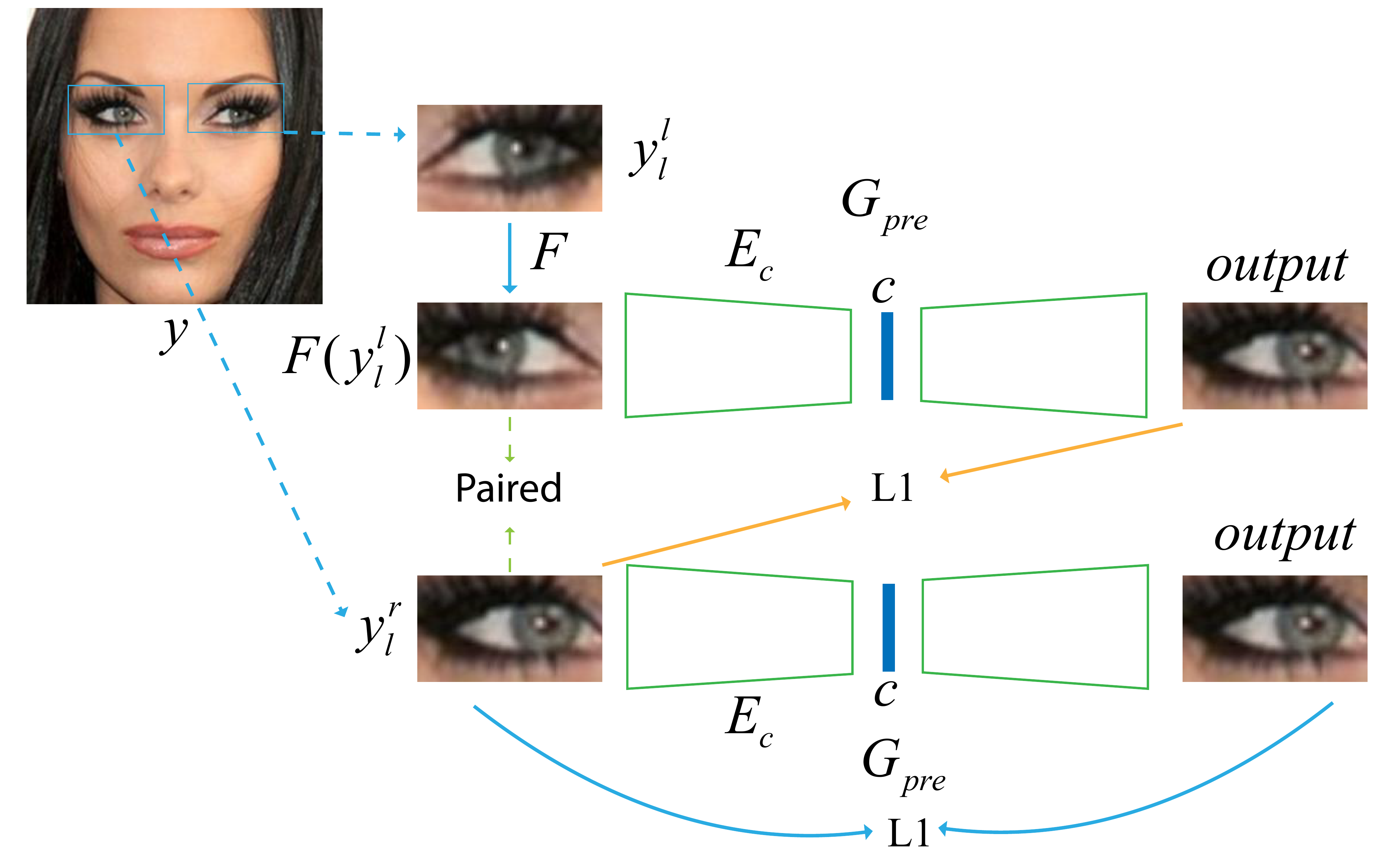}
\end{center}
\caption{The overview of pretrained autoencoder module with generator $G_{pre}$ (PAM). PAM is trained by a self-supervised learning strategy. In detail, we crop $y$ to both left eye $y^{l}_{l}$ and right eye $y^{r}_{l}$, then, flip $y^{l}_{l}$ by $F$ to attain the pairs $F(y^{l}_{l})$ and $y^{r}_{l}$ which have similar identity, but different gaze angle. Then, the pairs would be used to train $G_{pre}$ by reconstructing $y^{r}_{l}$.}
\label{fig:pre}
\end{figure}

Besides correcting the gaze to stare at the camera, a more general task is gaze animation, where the gaze direction should be modified. As shown in Fig.~\ref{fig:gam}, another generator $G_{y}$ is used to in-paint the face image without the eye region by performing the reconstruction learning. Moreover, we extend a new eye encoder $E_{r}$ for the eye region to extract the angle-specific feature, guiding the gaze redirection generation of the $G_{y}$. To achieve the disentanglement of the features, we propose a  Synthesis-As-Training method, in which we use the gaze-corrected generated images as training data for training GAM. In detail, our GAM is split into two stages. In the first stage (Left of Fig.~\ref{fig:gam}), we downsample $y^{h}$ to get $y$, and train the generator $G_{y}$ to fill-in the missing eye regions of an image ($y_{m} = M(y)$) and produce $\tilde y$. $G_{y}$ encodes the eye region with the latent code $r_{y} \in R^{128}$ by means of the encoder $E_{r}$. Moreover, $r_{y}$ is used as an extra input for the decoder in $G_{y}$. In this way, we can condition $G_{y}$ using the gaze-dependent feature $r_{y}$.
\begin{equation}
\begin{aligned}
r_{y} = E_{r}(M^{'}(y)), c_{y} = E_{c}(M^{'}(y)) \\
\tilde y = G_{y}(M(y), r_{y}, c_{y}).
\end{aligned}
\end{equation}

In the second stage (Right of Fig.~\ref{fig:gam}), we use GCM to correct the gaze of $y^{h}$, and it produces the synthetic sample $y^{h_{x}}$. Then, $y^{h_{x}}$ is downsampled to $y^{x}$ which is used for training $G_{y}$, just like $y$ does. With the paired samples $(y,y^{x})$, which have the same masked region $M(y)$ but different eye regions, we train GAM to ensure that the encoded feature from $E_{r}$ has a high correlation with the gaze angle:
\begin{equation}
\begin{aligned}
r_{y^{x}} = E_{r}(M^{'}(y^{x})), c_{y^{x}} = E_{c}(M^{'}(y^{x})) \\
\tilde y^{x} = G_{y}(M(y^{x}), r_{y^{x}}, c_{y^{x}}).
\end{aligned}
\end{equation}

Then, we can attain high-resolution results $\hat y^{h}$ and $\hat y^{h_{x}}$ by compensating for the high-frequency details using the laplacian reconstruction and LRM.

\subsection{Pretraining using Self-Supervised Learning} \label{PAM}

Preserving the consistency of the person's identity (e.g., the iris color, the eye shape) is difficult
with the inpainting-based method described. To mitigate this problem, we propose
to use the third generator $G_{pre}$, which is trained (PAM)
to learn a latent representation of the content features ($c$), conditioning both $G_{x}$ and $G_{y}$ to preserve the identity information of the generated results consistent with the input.

$G_{pre}$ is pre-trained using a self-supervised learning framework.
Although our training dataset is collected from the Internet, most images have a roughly frontal pose. As shown in Fig.~\ref{fig:pre}, we can easily collect paired eye-region images: The right eye $y^{r}_{l}$ is paired with the mirrored version $F(y^{l}_{l})$ of the left eye $y^{l}_{l}$.  
 Note that $y^{r}_{l}$ and  $F(y^{l}_{l})$
 have different gaze angles, but they have a similar eye shape and iris color. Because they belong to the same person. The same holds for
 $y^{l}_{l}$ and  $F(y^{r}_{l})$. 
 Note that the only information we need to collect these pairs is the eye region position (i.e., $M(x)$). At the same time, the mirroring operation ($F(\cdot)$) is a data augmentation technique commonly used in other self-supervised learning approaches.
We use these paired samples to pretrain $G_{pre}$ using the following objective function:

\begin{equation} \label{recon_model}
\begin{aligned}
\mathcal{L}_{pre} &=&  \Vert y^{l}_{l} - G_{pre}(y^{l}_{l}) \Vert_{1} 
+ \Vert y^{l}_{l} - G_{pre}(F(y^{r}_{l}) \Vert_{1} 
\\
&+& \Vert y^{r}_{l} - G_{pre}(y^{r}_{l}) \Vert_{1}
+ \Vert y^{r}_{l} - G_{pre}(F(y^{l}_{l})) \Vert_{1}.
\end{aligned}
\end{equation}

After training, the bottleneck features $c$ of $G_{pre}$ are almost angle-invariant, representing only the content information~(e.g., iris color, eye shape). Thus, we use the encoder network $E_{c}$ of $G_{pre}$ to provide extra input to $G_{x}$ and $G_{y}$ by means of its content features
(see Sec.~\ref{GCM} and \ref{GAM}). 
Conditioning the generation process of $G_{x}$ and $G_{y}$  using these content features, the inpainted results are more consistent with respect to the input identity information. 

\subsection{Loss Functions}

\textbf{Reconstruction Losses.}
We use a standard pixel-wise loss ($L1$) for training GCM. It is defined as:
\begin{equation} \label{recon_loss_x}
\begin{aligned}
\mathcal{L}^{x}_{re} = \Vert x - \tilde x \Vert_{1} + \Vert x^{h}_{l} - \hat x^{h}_{l} \Vert_{1},
\end{aligned}
\end{equation}
where $x^{h}_{l}$ and $\hat x^{h}_{l}$ are the eye regions of $x^{h}$ and $\hat x^{h}$, respectively.

And, the reconstruction loss for GAM is defined as:
\begin{equation} \label{recon_loss_y}
\begin{aligned}
\mathcal{L}^{y}_{re} = \Vert y - \tilde y \Vert_{1} + \Vert y^{h}_{l} - \hat y^{h}_{l} \Vert_{1}.
\end{aligned}
\end{equation}

The reconstruction loss of GAM for the synthesis-as-training method is defined as:
\begin{equation} \label{recon_loss_y_x}
\begin{aligned}
\mathcal{L}^{y^{x}}_{re} = \Vert y^{x} - \tilde y^{x} \Vert_{1} + \Vert y^{h_{x}}_{l} - \hat y^{h_{x}}_{l} \Vert_{1}.
\end{aligned}
\end{equation}

We use $\mathcal{L}^{y}_{re}$ + $\mathcal{L}^{y^{x}}_{re}$ as the objective function to train $G_{y}$.

\textbf{Global and Local Discriminators for Adversarial Learning.}
Since the $L1$ loss tends to produce blurry results \cite{isola2017image}, we use three different discriminators $D_{x}$, $D_{y}$ and $D_{h}$, adversarially trained together with $G_{x}$, $G_{y}$ and $G_{h}$, respectively. Moreover, inspired by~\cite{iizuka2017globally}, our discriminators $D_{x}$ and $D_{y}$ are composed of a global part, taking the whole face as input, and a local part with taking only the local eye region as input. The global part is used to coherent the entire image as a whole, while the local part makes the local region more realistic and sharper. We concatenate the final fully-connected feature maps of both parts, which are fed to a sigmoid function to predict the probability of the image being real. 

Different from $D_{x}$ and $D_{y}$, $D_{h}$  consists of only a local discriminator, which  uses the eye regions   $\hat x^{h}_{l}$ and $\hat y^{h}_{l}$ as fake inputs. In practice, we use crops slightly larger than the eye region as the input of the discriminator to alleviate the boundary mismatch problem.
The objective function of $D_{x}$ and $G_{x}$ is defined as:
\begin{eqnarray}
\mathop{min}\limits_{G_{x}}\mathop{max}\limits_{D_{x}}\mathcal{L}^{x}_{adv} &=& \mathbb{\mathbb{E}}_{x}[logD_{x}(x, M^{'}(x))]
\nonumber \\ &+& \mathbb{\mathbb{E}}_{\tilde x}[log(1-D_{x}(\tilde x, M^{'}(\tilde x)))]  \\ &+& \mathbb{\mathbb{E}}_{\tilde y^{x}}[log(1 - D_{x}(\tilde y^{x}, M^{'}(\tilde y^{x})))]. \nonumber
\label{eq_gan_loss_x}
\end{eqnarray}

The objective function of $D_{y}$ and $G_{y}$ is defined as:
\begin{eqnarray}
\mathop{min}\limits_{G_{y}}\mathop{max}\limits_{D_{y}}\mathcal{L}^{y}_{adv} &=& \mathbb{\mathbb{E}}_{y}[logD_{y}(y, M^{'}(y))] \nonumber \\ &+& \mathbb{\mathbb{E}}_{\tilde y}[log(1-D_{y}(\tilde y, M^{'}(\tilde y)))].
\label{eq_gan_loss_y}
\end{eqnarray}

Finally, the objective function of $D_{h}$ and $G_{h}$ is:
\begin{eqnarray}
\mathop{min}\limits_{G_{h}}\mathop{max}\limits_{D_{h}}\mathcal{L}^{h}_{adv} &=& \mathbb{\mathbb{E}}_{x^{h}}[logD_{h}(M^{'}(x^{h}))] \nonumber \\ &+& \mathbb{\mathbb{E}}_{\hat x^{h}_{l}}[log(1-D_{h}(\hat x^{h}_{l}))]
\nonumber \\ &+& \mathbb{\mathbb{E}}_{y^h}[logD_{h}(M^{'}(y^h))] \nonumber \\ &+& \mathbb{\mathbb{E}}_{\hat y^{h}_{l}}[log(1-D_{h}(\hat y^{h}_{l}))].
\label{eq_gan_loss_h}
\end{eqnarray}

\subsection{Overall Objective Function} 
Inspired by ~\cite{huang2018multimodal}, we use a latent-space reconstruction loss ($l_{fp}$) for the content features in the latent space to preserve further the identity information between the input image and the gaze-corrected result: \begin{equation} 
\begin{aligned}
\mathcal{L}_{fp} = \Vert c_{y} - E_{c}(M^{'}(\tilde y)) \Vert_{1} + \Vert c_{y^{x}} - E_{c}(M^{'}(\tilde y^{x})) \Vert_{1}.
\end{aligned}
\end{equation}

We use $- \ell^{x}_{adv}$, $- \ell^{y}_{adv}$ to train $D_{x}$ and $D_{y}$, respectively.
Concerning $G_{x}$, its overall loss is defined as:
\begin{eqnarray}
\mathcal{L}^{x}_{all} = \mathcal{L}^{x}_{adv} + \mathcal{L}^{h}_{adv} + \lambda_{1} \mathcal{L}^{x}_{re}.
\end{eqnarray}

For $G_{y}$ and $E_{r}$, the overall loss is defined as:
\begin{eqnarray}
\mathcal{L}^{y}_{all} &=& \mathcal{L}^{y}_{adv} + \mathcal{L}^{h}_{adv} + \lambda_{2} \mathcal{L}^{x}_{adv} \nonumber \\ &+&  \lambda_{3} \mathcal{L}^{y}_{re} + \lambda_{4} \mathcal{L}^{ y^{x}}_{re} + \lambda_{5} \mathcal{L}_{fp}.
\end{eqnarray}

$\lambda_{1}, \lambda_{2}, \lambda_{3}, \lambda_{4}$ and $\lambda_{5}$ are hyper-parameters controlling the contribution of each loss term. 
\subsection{Gaze Correction and Animation at Inference Time} \label{infer}

At inference time, given an image sample $y^{h}$,
we first downsample it to attain $y$, then obtain the gaze-corrected result $y^{x}$ using $G_{x}$, and finally output high-resolution results $y^{h_{x}}$ by using CFM, which can compensate for high-frequency texture details. In the case of gaze animation, as shown in Fig.~\ref{fig:gam}, we modify the angle representation $r$ by interpolating between $r_{y}$ and $r_{y^{x}}$, where both features correspond to the encoded angle features of the eye region $y_{l}$ and the eye region $y^{x}_{l}$, respectively. The interpolated angle representation can be fed to $G_{y}$ to obtain an intermediate result. We can produce high-resolution gaze animation results using CFM.

\section{Experiments}

\begin{figure*}[t]
\begin{center}
\includegraphics[width=1.0\linewidth]{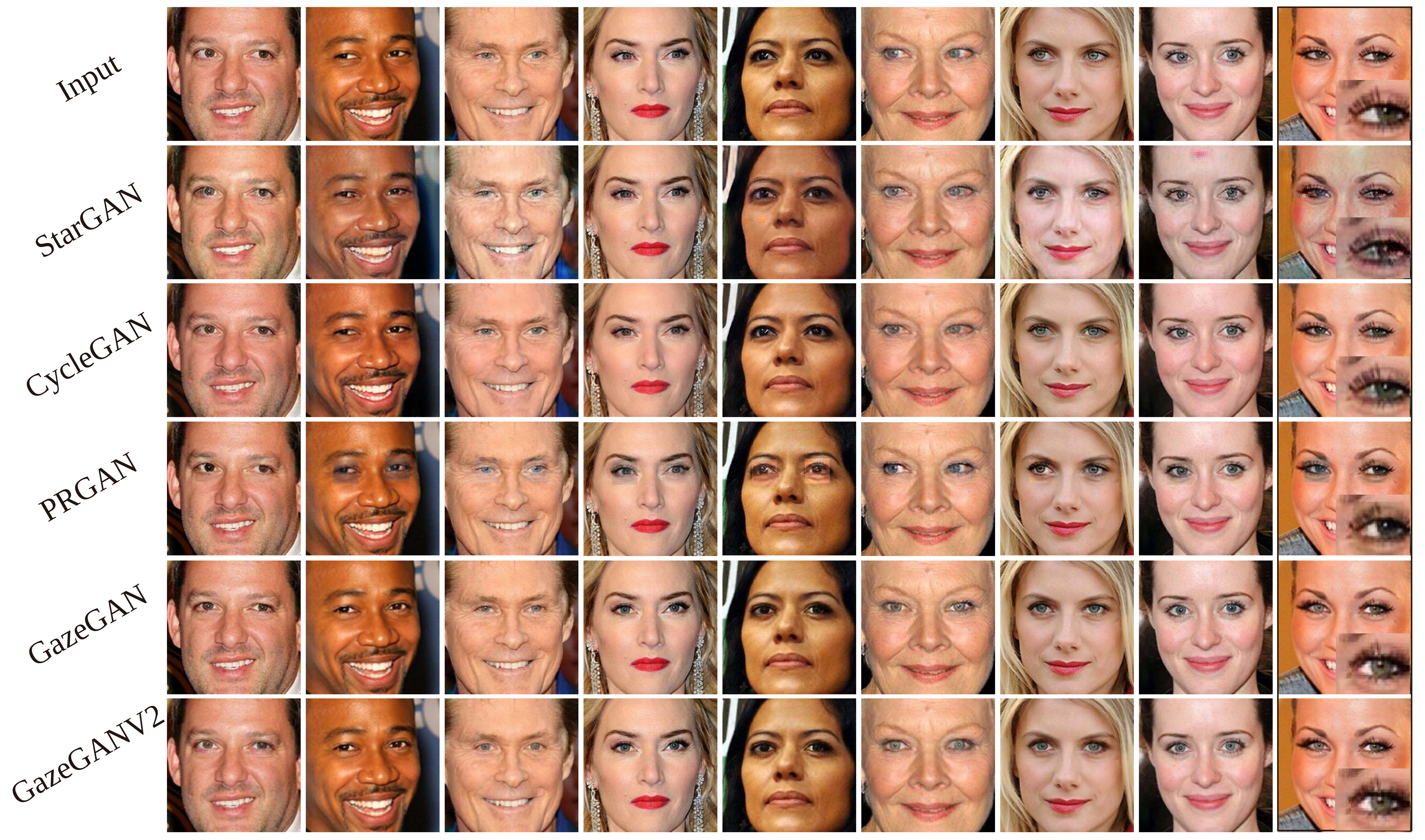}
\end{center}
\vspace{-0.3cm}
\caption{Qualitative comparison for the gaze correction task on the CelebGaze dataset. The first row shows the input images, and the following rows show the gaze correction results of StarGAN~\cite{Choi_2018_CVPR}, CycleGAN~\cite{Zhu_2017_ICCV}, PRGAN~\cite{he2019photo}, GazeGAN and GazeGANV2. Magnified left eyes are shown in the last column. Zoom in for the best of view.}
\label{fig:exp1}
\end{figure*}

\begin{figure*}[t]
\begin{center}
\includegraphics[width=1.0\linewidth]{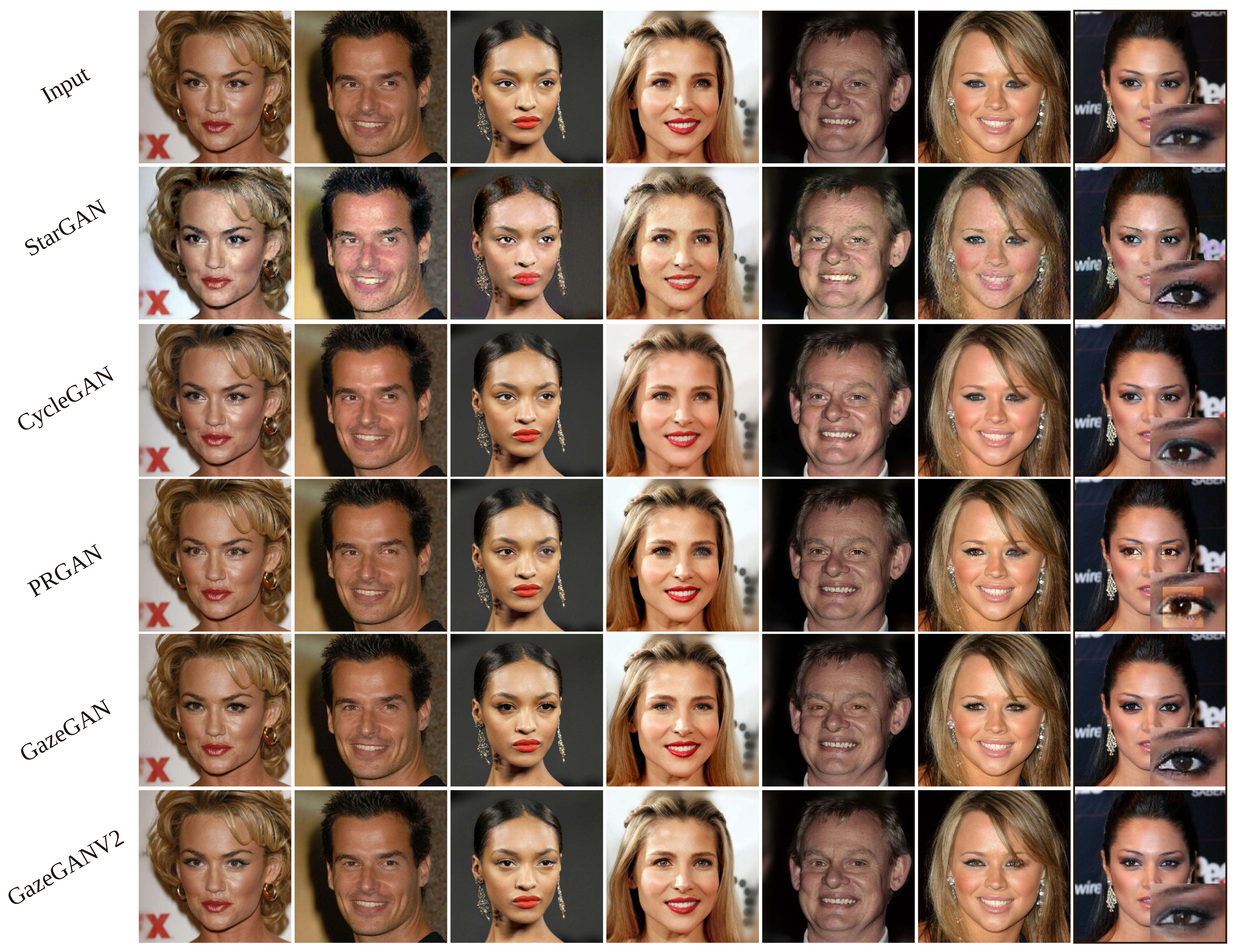}
\end{center}
\vspace{-0.5cm}
\caption{Qualitative comparison for the gaze correction task on CelebHQGaze dataset. The first row shows the input images, and the following rows show the gaze correction results of StarGAN~\cite{Choi_2018_CVPR}, CycleGAN~\cite{Zhu_2017_ICCV}, PRGAN~\cite{he2019photo}, GazeGAN and GazeGANV2. Magnified left eyes are shown in the last column. Zoom in for the best of view.}
\vspace{-0.4cm}
\label{fig:exp_hq1}
\end{figure*}

This section introduces the details of our datasets, our network training, and baseline models. 
Then, we compare the proposed method with the state-of-the-art methods of gaze correction in the wild using both qualitative and quantitative evaluations. 
Next, we demonstrate the effectiveness of the proposed method on gaze animation with various outputs by interpolating and extrapolating in the latent space. 
Finally, we present detailed ablation studies to validate the effect of each component of our model, i.e., the Synthesis-As-Training method, the Pretrained Autoencoder (PAM) with self-supervised mirror learning, the Latent Reconstruction Loss, and the Coarse-to-Fine Module (CFM). For brevity, we refer to the method presented in \cite{10.1145/3394171.3413981} as~\textbf{GazeGAN} and the extended version introduced in this paper  as~\textbf{GazeGANV2}. Note that we do not use any post-processing step for GazeGAN and GazeGANV2. 

\subsection{Datasets}
Most of the existing benchmarks~\cite{funes2014eyediap,smith2013gaze,zhang2017mpiigaze, zhang2020eth} do not contain enough image variability (e.g., a wide gaze-direction range, various head poses, and different illumination conditions) for our gaze correction task in the wild. Recently,~\cite{kellnhofer2019gaze360} presented a large-scale gaze tracking dataset, called Gaze360, for robust 3D gaze estimation in unconstrained images. Although this dataset has been labeled with a 3D gaze direction with a wide range of angles and head poses, it still lacks high-resolution images for face and eye regions. Moreover, this dataset does not provide annotations of the eye gaze staring at the camera, which is required in our domain set $X$. More recently, ~\cite{Zhang2020ETHXGaze} proposed a large scale~(over 1 million samples) of high-resolution images for gaze estimation. However, these images are collected in laboratory conditions and are not suitable for our gaze correction task in the wild. To remedy this problem, we propose collecting new datasets consisting of lots of high-resolution portraits without labelling head poses and gaze information. In detail, five volunteers are asked to divide the row data (face) into two domains according to whether the face eyes are staring at the camera. The gaze and head estimation model can automate ‘Staring at the camera’ annotation. However, the existing state-of-the-art methods~\cite{kellnhofer2019gaze360,cai2021gaze} cannot achieve accurate gaze estimation for CelebHQGaze, as an overlarge domain shift exists between training data and test data. More details about our datasets can be found below.

{\bfseries CelebGaze.}
CelebGaze consists of 25,283  celebrity images, most of which have been collected from CelebA~\cite{liu2015deep} and a minority from the Internet. Specifically, there are 21,832 face images with the eyes staring at the camera and 3,451 face images with the eyes looking somewhere else. We crop all the images to $256 \times 256$ and compute the eye mask region using Dlib~\cite{king2009dlib}. Specifically, we use Dlib to extract 68 facial landmarks, and we compute the mean of 6 points near the eye region, which is the center point of the mask. The size of the mask is fixed to $30\times 50$. We randomly select 300 samples from domain $Y$ and 100 samples from domain $X$ as their corresponding test sets, and we use the remaining images for the training set. Note that this dataset is unpaired and not labeled with the specific eye angle or the head pose information. We show some samples of the CelebGaze dataset in Fig.~\ref{fig:newgaze}.

\begin{figure}[t]
\begin{center}
\vspace{-0.3cm}
\includegraphics[width=1.0\linewidth]{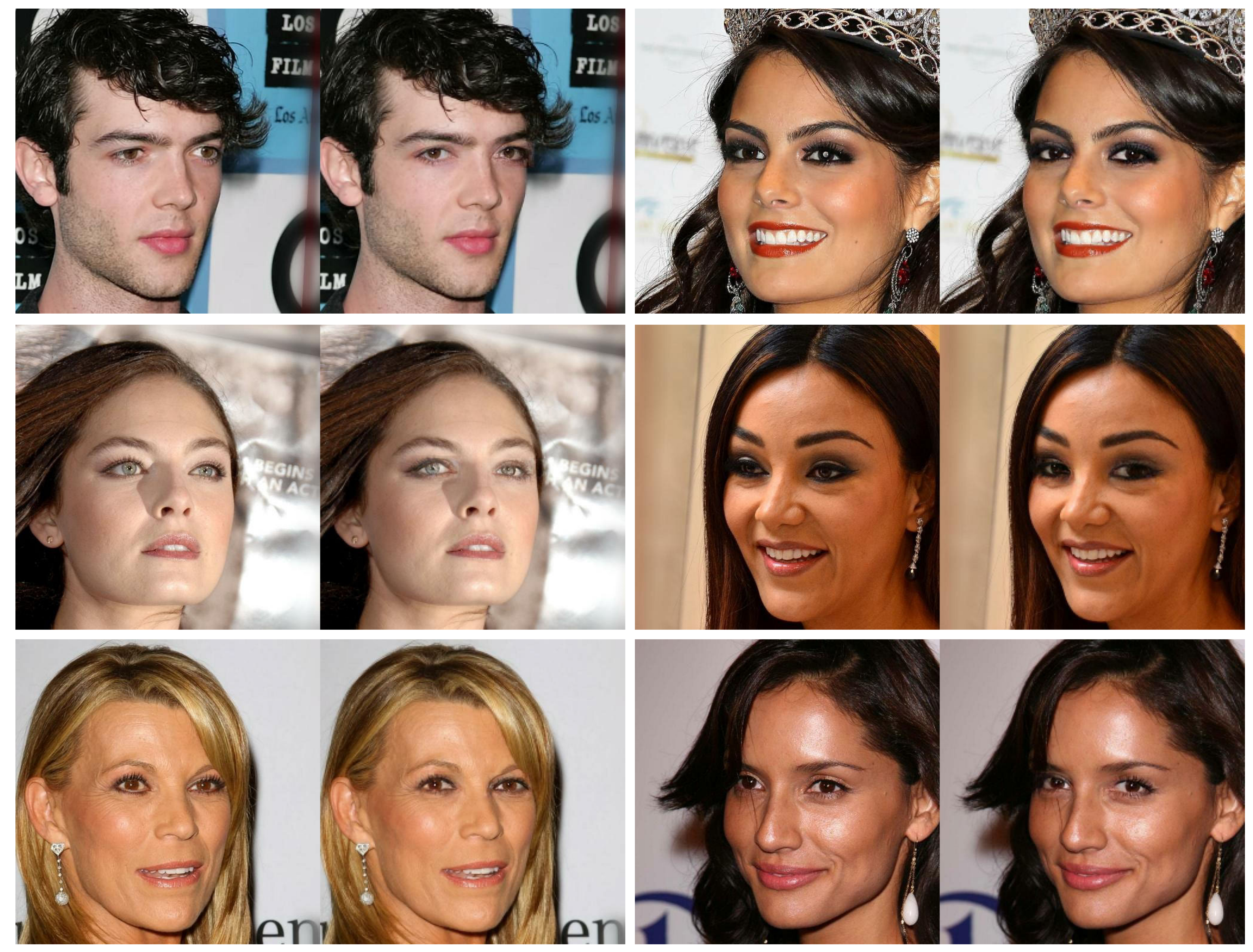}
\end{center}
\vspace{-0.3cm}
\caption{More gaze correction results to show that our model can handle diverse head poses.}
\vspace{-0.2cm}
\label{fig:exp_hq2}
\end{figure}

{\bfseries CelebHQGaze.}
CelebHQGaze consists of 29,255 high-resolution celebrity images that are collected from CelebA-HQ~\cite{lee2020maskgan}. It consists of 21,005 face images with the eyes staring at the camera and 8,250 face images with eyes looking somewhere else. Similarly to CelebGaze, we extract facial landmarks and generate the mask. All images are cropped to $512 \times 512$, and the mask size is fixed to $46 \times 80$. Similar to the CelebGaze dataset, also for the CelebHQGaze, we randomly select 300 samples from domain $Y$ and 100 samples from domain $X$ for the test set, and we use all the remaining images for the training set. We show two samples of the CelebHQGaze dataset in Fig.~\ref{fig:newgaze}. 

\begin{figure*}[t]
\begin{center}
\includegraphics[width=1.0\linewidth]{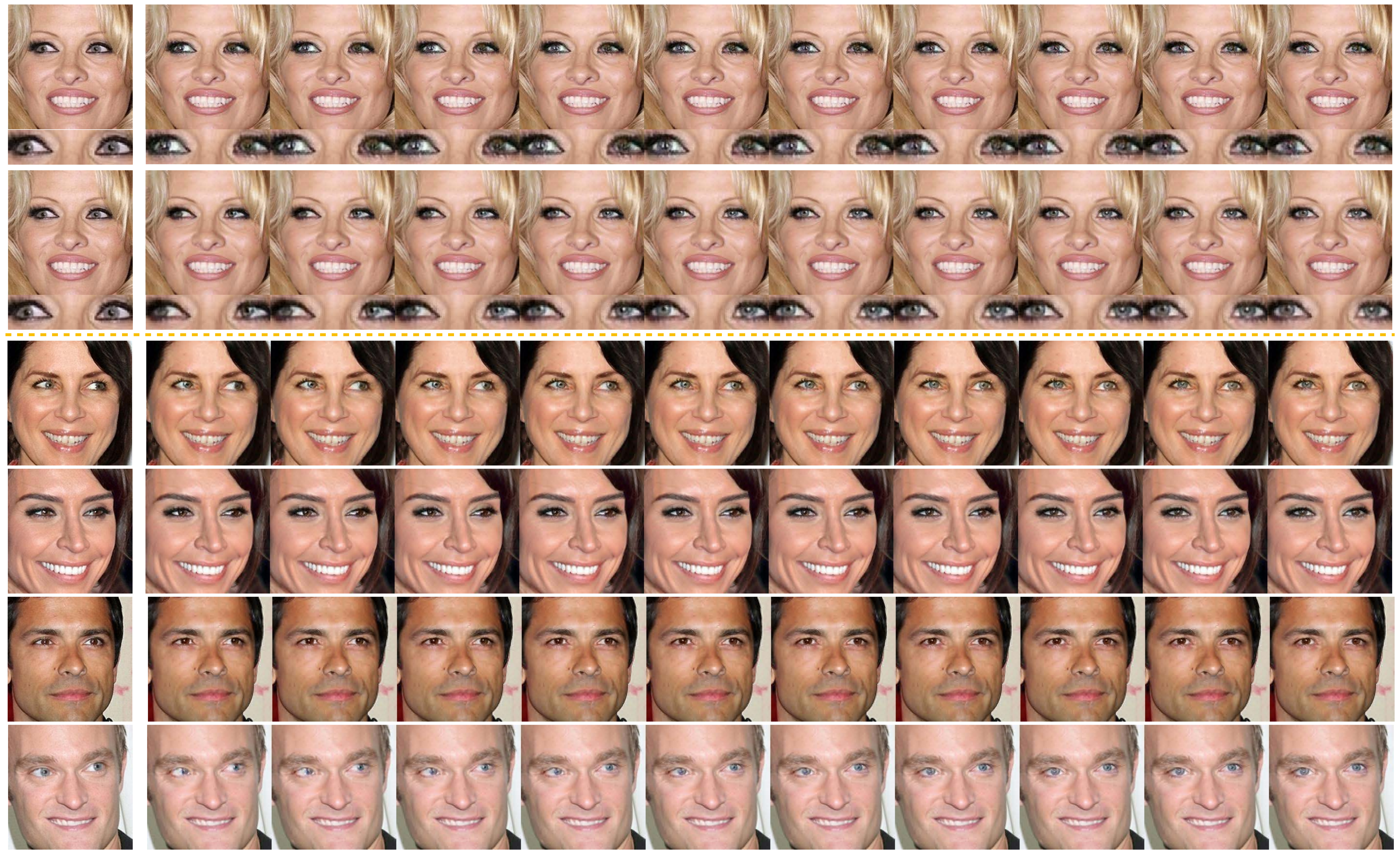}
\end{center}
\vspace{-0.3cm}
\caption{Gaze animation results using the interpolation of the latent features $r$ on the CelebGaze dataset. The top two rows show the images generated by GazeGAN and GazeGANV2, respectively,
jointly with the eye regions. The other rows show the gaze animation results of GazeGANV2.
The first and the last columns show the input images and the gaze-corrected results, respectively. The middle columns show the interpolated images.}
\label{fig:exp2_1}
\vspace{-0.2cm}
\end{figure*}

\subsection{Training Details}
We first train the PAM module. Then, the discriminators  $D_{x}$, $D_{y}$ and $D_{h}$ and the generators $G_{x}$ and $G_{y}$ and $G_{h}$ are jointly optimized. We use the Adam optimizer with $\beta_{1}=0.5$ and $\beta_{2}=0.999$. The batch size is 16 for CelebGaze and 8 for CelebHQGaze. The initial learning rate is 0.0005 for PAM, 0.0004 for $G_{h}$, and 0.0001 for the three discriminators and the two generators in the first 20,000 iterations. The learning rate is linearly decayed to 0 over the remaining iterations. The loss coefficients $\lambda_{1}, \lambda_{2}, \lambda_{3}, \lambda_{4}$ are all set to 1, while $\lambda_{5}$ is 0.1. To stabilize the network training in the adversarial learning, we use spectral normalization~\cite{miyato2018spectral} for all the conv-layers of the three discriminators, but not for the generators. Our method is implemented in Tensorflow and trained with a single NVIDIA Titan X GPU.

\subsection{Baseline Models}
{\bfseries Gaze Correction.}
PRGAN~\cite{he2019photo} achieved state-of-the-art gaze redirection results on the Columbia gaze dataset~\cite{smith2013gaze} based on a single encoder-decoder network with adversarial learning, similarly to the  StarGAN architecture \cite{Choi_2018_CVPR}. The original PRGAN is trained on paired samples with labeled angles. To train PRGAN on the proposed CelebGaze and CelebHQGaze datasets, we remove the VGG perceptual loss of PRGAN, and learn the gaze redirection task between domain $X$ and $Y$. We train PRGAN only with the local eye region, the same way as the original paper.

\begin{figure*}[t]
\begin{center}
\includegraphics[width=1.0\linewidth]{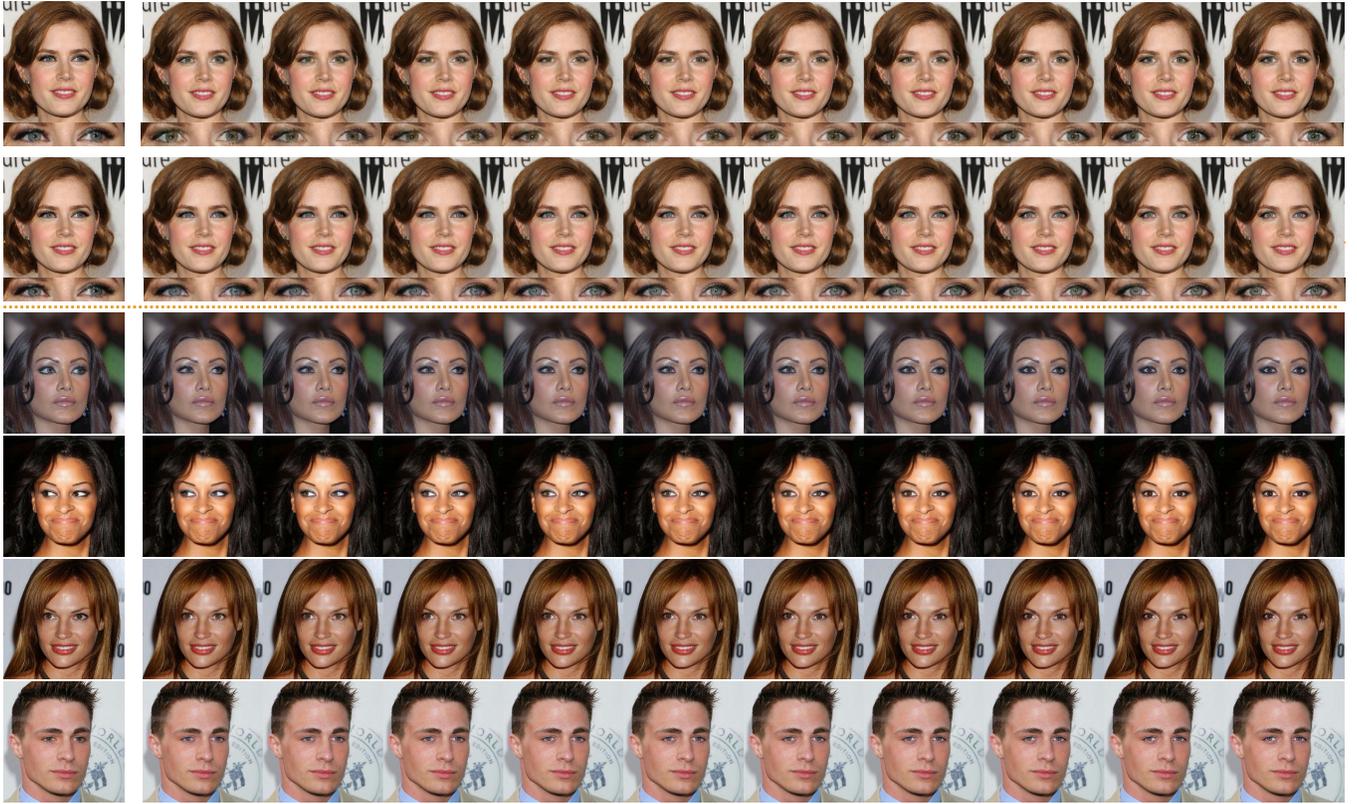}
\end{center}
\vspace{-0.3cm}
\caption{Gaze animation results using the interpolation of the latent features $r$ on the CelebHQGaze dataset. The top two rows show the images generated by GazeGAN and GazeGANV2, respectively, jointly with the eye regions. The other rows show the gaze animation results of GazeGANV2. The first and the last columns show the input images and gaze-corrected results, respectively. The middle columns show the interpolated images.}
\label{fig:exp2_2}
\vspace{-0.2cm}
\end{figure*}

\begin{figure}[t]
\begin{center}
\includegraphics[width=1.0\linewidth]{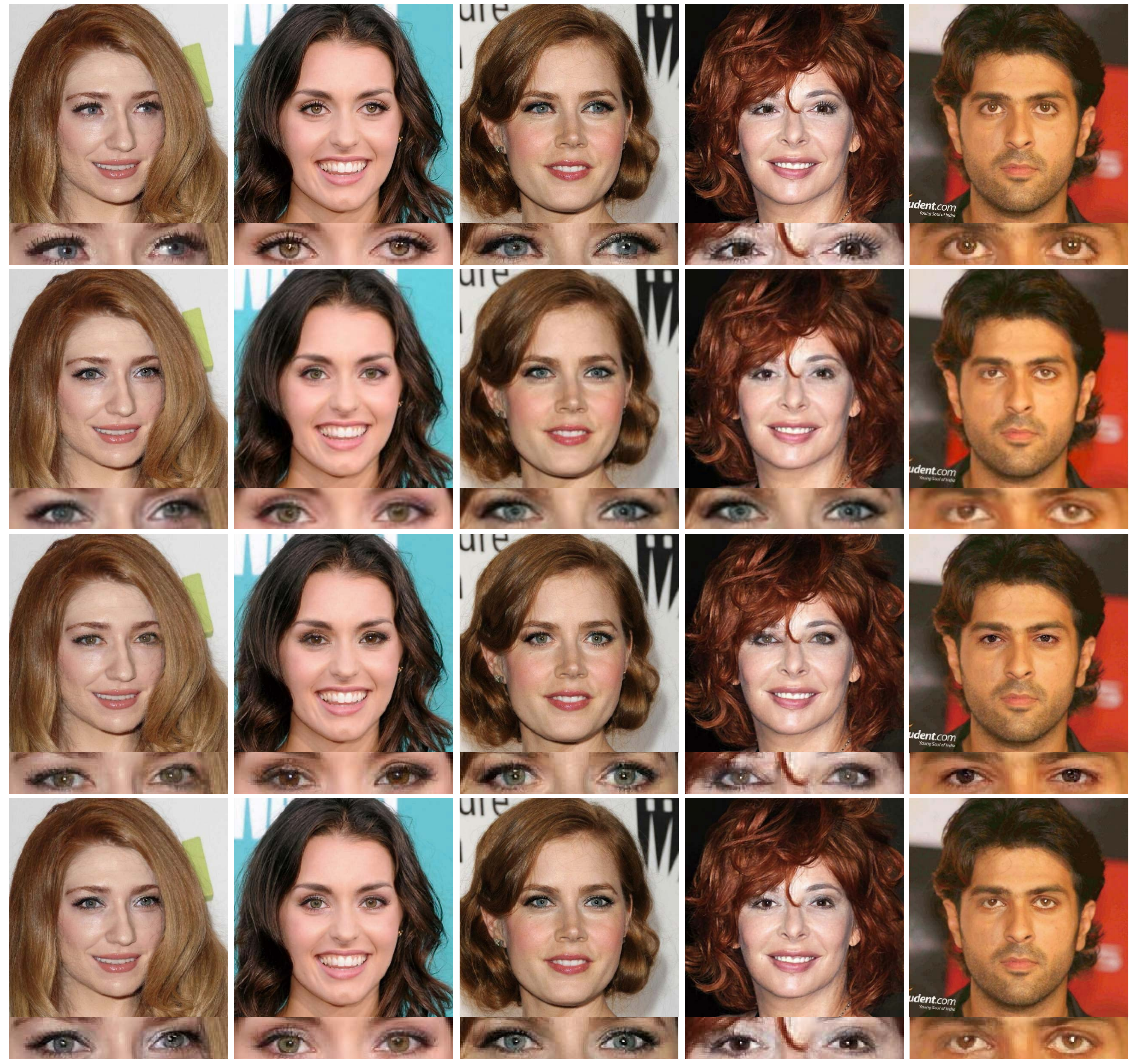}
\end{center}
\vspace{-0.3cm}
\caption{A qualitative comparison between GazeGANV2~(4th row), GazeGANV2 W/O $A$~(3rd row), and GazeGANV2 W/O $D$~(2nd row). The first row shows the input images, and the bottom of every row shows the zoom-in eye regions.}
\vspace{-0.2cm}
\label{fig:exp5}
\end{figure}

\begin{table*}[t]
\caption{Quantitative results on both the CelebGaze and the CelebHQGaze dataset. The higher is better for MSSSIM and the user study; the lower is better for LPIPS and FID. The columns Params and FPS report the corresponding network parameters and frame per second at test time, respectively. US: user studies.}
\label{tab:Quan_evaluation1}
\centering
	\resizebox{1\linewidth}{!}{
\begin{tabular}{lllllllllllll}
\toprule
\multirow{2}{*}{Method} & \multicolumn{6}{c}{CelebGaze} & \multicolumn{6}{c}{CelebHQGaze} \\
\cmidrule(r){2-7}
\cmidrule(r){8-13}
 & MSSSIM $\uparrow$ & LPIPS $\downarrow$  & FID $\downarrow$  & US $\uparrow$ & Params $\downarrow$ & FPS $\downarrow$  & MSSSIM $\uparrow$ & LPIPS $\downarrow$  & FID $\downarrow$  & US $\uparrow$ & Params $\downarrow$ & FPS $\downarrow$ \\
\midrule
Other & - & - & - & 24.20\% & - & - & -  & - & - & 23.20\% & - & - \\
StarGAN~\cite{Choi_2018_CVPR} & 0.96 & 0.073 & 82.49 & 3.400\% & - & - & 0.94 & 0.084 & 185.47 & 4.400\% & - & - \\
%StarGANA & 0.998 & 0.0022 & 6.66\% & {\bfseries 50.83} & {\bfseries 28.05M} & 0.997 & 0.0024 & 3.20\%  & 89.28 & 28.25M \\
CycleGAN~\cite{Zhu_2017_ICCV} & 0.99 & 0.026 & 70.12 & 15.00\% & - & - & 0.98 & 0.028 & {\bfseries 53.690} & 8.670\% & - & - \\
PRGAN~\cite{he2019photo} & 1.00 & 0.000 & 84.61 & 8.330\% & - & - & 1.00 & 0.000 & 106.79 & 22.40\% &  - & - \\
GazeGAN & {\bfseries 1.00} & {\bfseries 0.000} & 62.12 & 22.40\% & 73.26M & 30.29 &   {\bfseries 1.00}  & {\bfseries 0.000} & 60.520 & 25.50\% & 183.2M & 23.20 \\
GazeGANV2 & {\bfseries 1.00} & {\bfseries 0.000} & {\bfseries 56.37} & {\bfseries 32.40\%} & {\bfseries 47.20M} & {\bfseries 38.40} & {\bfseries 1.00} & {\bfseries 0.000} & 63.590 & {\bfseries 27.30\%} & {\bfseries 84.18M} & {\bfseries 27.70} \\
\hline
GT & 1.00 & 0.000 & - & 100\% & - & - & 1.00 & 0.000 & -  & 100\% & - & - \\
\bottomrule
\end{tabular}}
\end{table*}

\begin{figure*}[t]
\begin{center}
\includegraphics[width=1.0\linewidth]{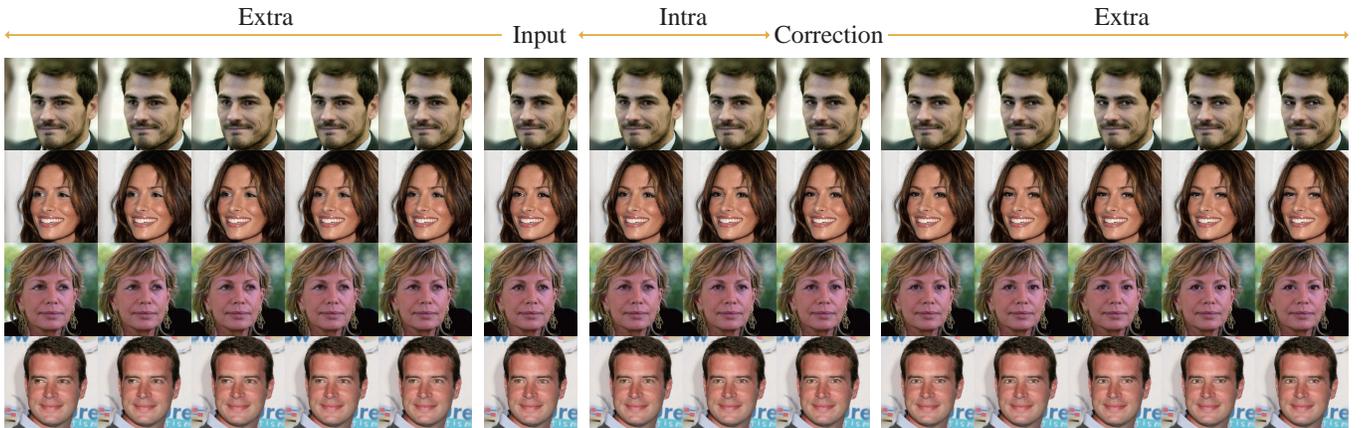}
\end{center}
\vspace{-0.5cm}
\caption{Gaze animation examples are obtained by both interpolation and extrapolation of the latent features $r$. Extra: extrapolation; Intra: interpolation.}
\label{fig:exp2_extra}
\end{figure*}

{\bfseries Facial Attribute Manipulation.}
Gaze correction and animation can be regarded as a sub-task of facial attribute manipulation. Recently, StarGAN~\cite{Choi_2018_CVPR}  achieved very high-quality results in facial attribute manipulation. We train StarGAN on the CelebGaze dataset to learn the translation mapping between domain $X$ and domain $Y$. 

Moreover, gaze correction can be considered as an image translation task. Thus, we adopt CycleGAN as another baseline for our experiments. Note that we do not compare GazeGAN with AttGAN~\cite{he2019attgan}, STGAN~\cite{liu2019stgan}, RelGAN~\cite{wu2019relgan}, CAFE-GAN~\cite{gicafe2020}, SSCGAN~\cite{chusscgan2020} as they have a performance very close to StarGAN in the facial attribute manipulation task. We use the public code of StarGAN~\footnote{\url{https://github.com/yunjey/StarGAN}}, CycleGAN~\footnote{\url{https://github.com/junyanz/pytorch-CycleGAN-and-pix2pix}} and PRGAN~\footnote{\url{https://github.com/HzDmS/gaze\_redirection}}.

\subsection{Gaze Correction}

This section qualitatively and quantitatively compares the proposed method with state-of-the-arts on both CelebGaze and CelebHQGaze datasets for the gaze correction task.

{\bfseries Qualitative Results.}
As shown in the last row of Fig.~\ref{fig:exp1} and Fig.~\ref{fig:exp_hq1}, GazeGANV2 can correct the eyes to look at the camera while preserving the identity information such as the eye shape and the iris color, validating the effectiveness of the proposed method. The 2nd row of the figure shows the results of StarGAN~\cite{Choi_2018_CVPR}. We note that StarGAN could not produce precise gazes staring at the camera, and it suffers from a low-quality generation with lots of artifacts (Zoom in for the best of view). The results of CycleGAN are shown in the 3th row. Although the results of CycleGAN are very realistic and with few artifacts in the eye region, this method does not produce a precise correction of the gaze direction (e.g., see the magnified eye regions of Fig.~\ref{fig:exp1} and Fig.~\ref{fig:exp_hq1}). We explain that both StarGAN and CycleGAN use the cycle-consistency loss, which requires that the mapping between $X$ and $Y$ be continuous and invertible. According to the invariance of the Domain Theorem\footnote{\url{https://en.wikipedia.org/wiki/Invariance\_of\_domain}}, the intrinsic dimensions of the two domains should be the same. However, the intrinsic dimension of $Y$ is much larger than $X$, as $Y$ has more variations for the gaze angle than $X$. 

Moreover, we compare GazeGANV2 with PRGAN~\cite{he2019photo}. PRGAN is trained using only local eye regions (same as in the original paper), which may help focus on the translation of the eye region. The results of PRGAN are shown in the 4th row of Fig.~\ref{fig:exp1}. Compared with GazeGANV2, PRGAN does not produce precise and realistic correction results. Additionally, PRGAN suffers from the boundary mismatch problem between the local eye region and the global face~(see the last column of Fig.~\ref{fig:exp_hq1}). 

Finally, as shown in the last rows of both Fig.~\ref{fig:exp1} and Fig.~\ref{fig:exp_hq1}, comparing GazeGANV2 with GazeGAN, we observe that both models can produce realistic and faithfully results. Additionally, we show more results of portraits with a diverse head pose. Fig.~\ref{fig:exp_hq2} shows that our model can achieve acceptable gaze-correction results for portraits with different head poses.

{\bfseries Quantitative Evaluation Protocol.}
The qualitative evaluation has validated the effectiveness and the superiority of our proposed GazeGANV2 in the gaze correction task. To further support the previous evaluation with quantitative results, we use the MSSSIM~\cite{wang2003multiscale} and the LPIPS~\cite{zhang2018perceptual} metrics to measure the preservation ability of the {\it irrelevant regions}, i.e.,, the whole image except the eye region ($M(y^{h})$). Specifically, we compute the mean MSSSIM and LPIPS scores between $M(y^{h})$ and $M(\hat y^{h})$ across all the {\it test} data of  $Y^{h}$. Moreover, the Fr\'echet Inception Distance (FID)~\cite{DBLP:journals/corr/HeuselRUNKH17} has been shown to correlate well with the human judgment and has become a popular metric for GAN-based methods. We use it to evaluate the quality of the generated eye region for the gaze correction and the gaze animation tasks.  

In addition to the aforementioned automatic metrics, we conduct a user study to compare the results of the gaze correction task of different models. In detail, given an input face image of the CelebGaze or CelebHQGaze test dataset (extracted from $Y$), we show the gaze-corrected results produced by different models to 30 respondents, who were
asked to select the best image based on the perceptual realism and the precision of the gaze correction. They also can select ``Other'', which means that the results of all the models are not satisfactory enough. This study is based on 50 questions (i.e., 50 randomly sampled images) for each respondent.

{\bfseries Quantitative Results.}
The first two columns of the left part~(CelebGaze) and the right part~(CelebHQGaze) of Table~\ref{tab:Quan_evaluation1} show the MSSSIM and LPIPS scores evaluating the preservation ability of the corrected images using different methods. GazeGANV2 and PRGAN obtain the best results, with 1.0 for MSSSIM and 0.0 for LPIPS. The original irrelevant regions are integrated with the generated eye region in both models using binary masks. StarGAN and CycleGAN get the worse irrelevant region preservation scores. The FID scores of the eye regions are reported in the 3rd column. In the CelebGaze dataset, GazeGANV2 and GazeGAN outperform all the other methods, reaching comparable scores on the CelebHQGaze dataset. Though CycleGAN has the best FID scores, it fails to generate precise gaze correction results. The penultimate column of both parts in Table~\ref{tab:Quan_evaluation1} shows the evaluation results of the user study. For the CelebGaze dataset, the average vote for GazeGANV2 is $32.40\%$, which is higher than all the other methods, i.e., $3.40\%$ for StarGAN, $15.00\%$ for CycleGAN, $8.33\%$ for PRGAN. The same conclusions can be drawn with the CelebHQGaze dataset. Importantly, GazeGANV2 achieves a performance very close to GazeGAN. However, it has fewer parameters and higher FPS, as shown in the last two columns of Table~\ref{tab:Quan_evaluation1}.

Overall, the qualitative and quantitative evaluations demonstrate the effectiveness and superiority of the proposed approach.

\begin{table} \small
\caption{A comparison on the gaze animation task between GazeGAN and GazeGANV2 with respect to the generation quality.}
\label{tab:exp2_FID}
\centering
	\resizebox{1\linewidth}{!}{
\begin{tabular}{lcccc}
\toprule
\multirow{2}{*}{Method} & \multicolumn{2}{c}{CelebGaze} & \multicolumn{2}{c}{CelebHQGaze} \\
\cmidrule(r){2-3}
\cmidrule(r){4-5}
 & GazeGAN & GazeGANV2 & GazeGAN & GazeGANV2 \\
\midrule
FID $\downarrow$  & 80.31 & {\bfseries 53.32} & {\bfseries 70.56} & 71.37 \\
\bottomrule
\end{tabular}}
\end{table}

\subsection{Gaze Animation}
The bottom of Fig.~\ref{fig:exp2_1} and~\ref{fig:exp2_2} show gaze animation results using input images with various gaze directions. The latent-space interpolation results are smooth and plausible in each row. Each column has a different gaze direction angle, but the identity information is overall preserved (e.g., the eye shape, the iris color, etc.). 
 
The top rows of Fig.~\ref{fig:exp2_1} and~\ref{fig:exp2_2} show a gaze animation comparison between GazeGAN and GazeGANV2. GazeGANV2 can produce more realistic images with fewer artifacts than GazeGAN on the CelebGaze dataset, while they have comparable performance on the CelebHQGaze dataset. The quantitative result confirms it in Table~\ref{tab:exp2_FID}.
 
Finally, we show gaze animation results obtained by {\it extrapolating} the features $r$, in addition to using interpolation methods. With ``extrapolation," we mean that we use values of $r$ which lie in the line connecting  $r_{y}$ with $r_{y^{x}}$, but they are outside these two points.
As shown in Fig.~\ref{fig:exp2_extra}, our method not only achieves high-quality interpolation results but is also able to produce plausible gaze animations when the gaze angles are outside the range between the input and the gaze-corrected output. 
 
\subsection{Ablation Study}
\label{AblationStudy}

In this section, we conduct extensive ablation studies to investigate the contribution of each of four critical components of our proposed GazeGANV2, i.e., the Pretrained Autoencoder for content feature extraction, the Synthesis-As-Training method, the Latent Reconstruction Loss $\mathcal{L}_{fp}$ and the Coarse-to-Fine Module. We refer to these components as $A$, $B$, $C$, and $D$, respectively. 

{\bfseries Pretrained Autoencoder (PAM).} Sec.~\ref{PAM} shows how self-supervised learning is used to pretrain a content encoder. This encoder produces identity-specific features which condition the generation process of $G_{x}$ and $G_{y}$. Here we analyze this aspect of our method.

\begin{table}[!t] \small
\centering
\caption{Comparison between GazeGANV2 and GazeGANV2 W/O $A$, where the latter denotes removing the content representation extracted from~$E_{c}$. The scores are measured between the input image $x$ and inpainted result $\tilde x$ across the test data from $X$. Note that evaluation samples are from CelebHQGaze.}
\begin{tabular}{ccc}
\toprule
Metrics &  GazeGANV2 & GazeGANV2 W/O A \\
\midrule
MSSSIM $\uparrow$  & {\bfseries 0.6080} & 0.5230 \\
LPIPS $\downarrow$  & {\bfseries 0.1680} & 0.2646 \\    
\bottomrule
\end{tabular}
\label{tab:exp2}
\vspace{-0.3cm}
\end{table}

\begin{table}[t] \small
\centering
\caption{Comparison with GazeGAN W/O $C$, which denotes removing the latent reconstruction loss $\mathcal{L}_{fp}$. The scores are measured between the input image $y$ and the reconstruction result $\tilde y$ across all the test data of domain $Y$. The evaluation is based on the CelebHQGaze dataset. 
}
\begin{tabular}{ccc}
\toprule
Metrics &  GazeGANV2 & GazeGANV2 W/O $C$ \\
\midrule
MSSSIM $\uparrow$  & {\bfseries 0.6290} & 0.6100 \\
LPIPS $\downarrow$  & {\bfseries 0.2328} & 0.2372 \\   
\bottomrule  
\end{tabular}
\label{tab:AB_C}
\vspace{-0.4cm}
\end{table}

\begin{figure}[t]
\begin{center}
\includegraphics[width=1.0\linewidth]{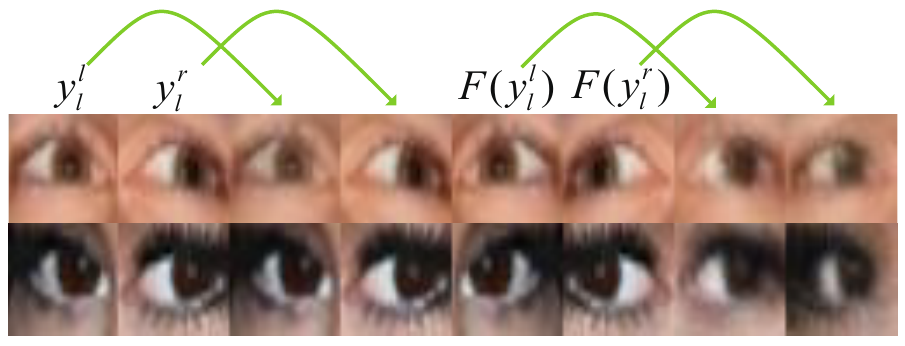}
\end{center}
\vspace{-0.3cm}
\caption{The visualization for the results of PAM with taking $y^{l}_{l}$, $y^{r}_{l}$, $F(y^{l}_{l})$ and $F(y^{r}_{l})$ as inputs, respectively. The arrows point to the generated sample.}
\vspace{-0.2cm}
\label{fig:AB_A2}
\end{figure}

\begin{figure}
\begin{center}
\includegraphics[width=1.0\linewidth]{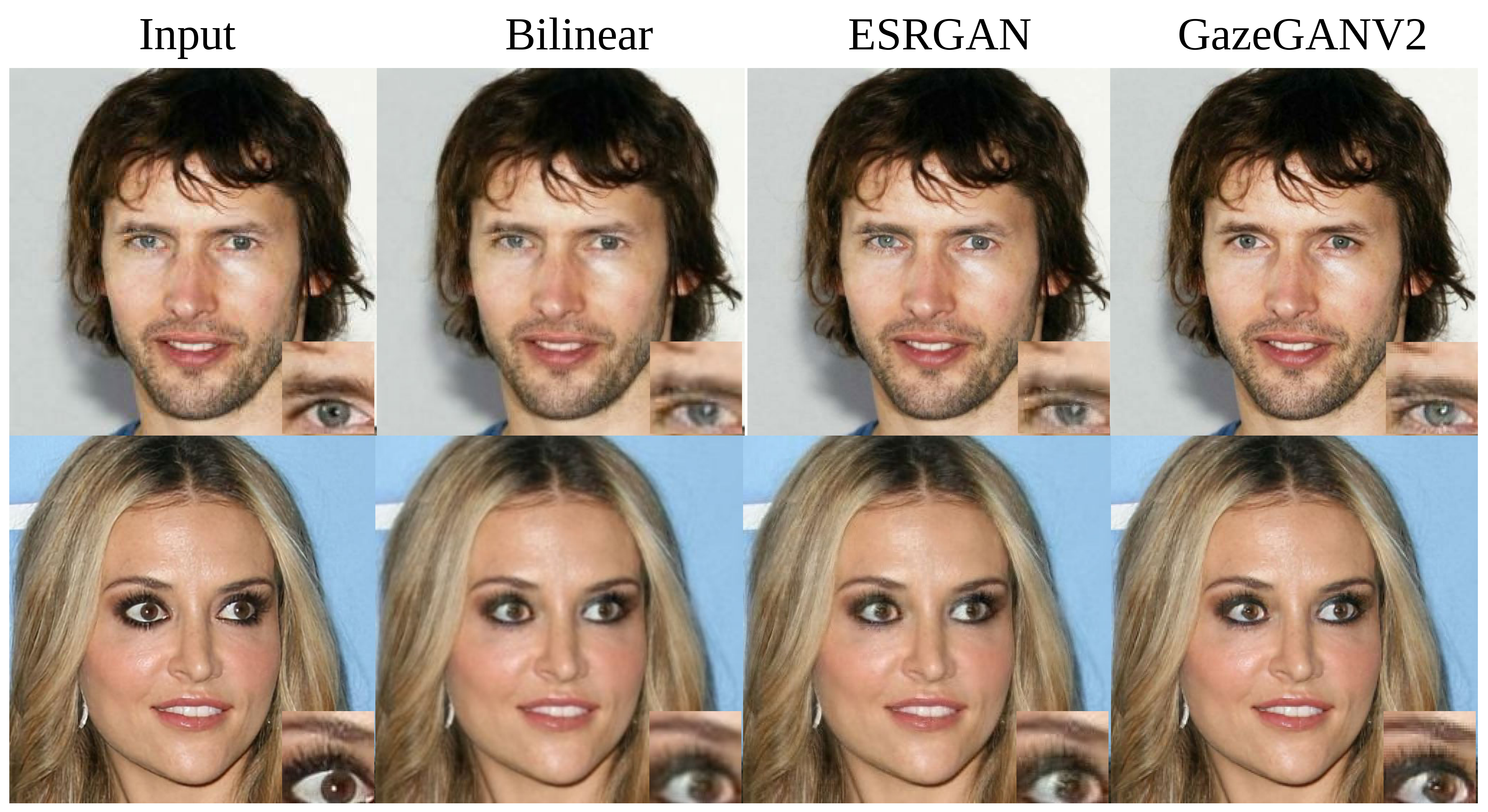}
\end{center}
\vspace{-0.3cm}
\caption{Qualitative comparison between GazeGANV2 with Bilinear and super-resolution model ESRGAN~\cite{wang2018esrgan}. Note that the input image would be downsampled 2$\times$ as input of the inpainting model.}
\vspace{-0.2cm}
\label{fig:AB_D3}
\end{figure}

\begin{figure}[t]
\begin{center}
\includegraphics[width=1.0\linewidth]{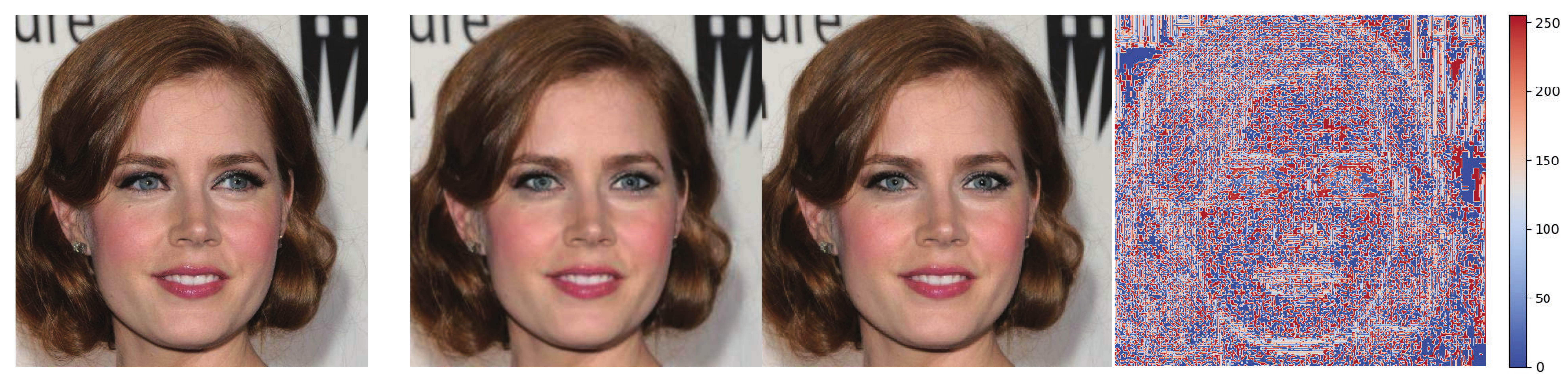}
\end{center}
\vspace{-0.4cm}
\caption{A qualitative comparison between the gaze-correction results produced by GazeGANV2~(3rd column) and GazeGANV2 W/O $D$~(2nd column). The 1st column shows the input image, and the final column is a heatmap of the difference between the 3rd column and 2nd column. This residual image clearly shows semantic and texture information.}
\vspace{-0.3cm}
\label{fig:AB_D2}
\end{figure}

\begin{figure}
\begin{center}
\includegraphics[width=1.0\linewidth]{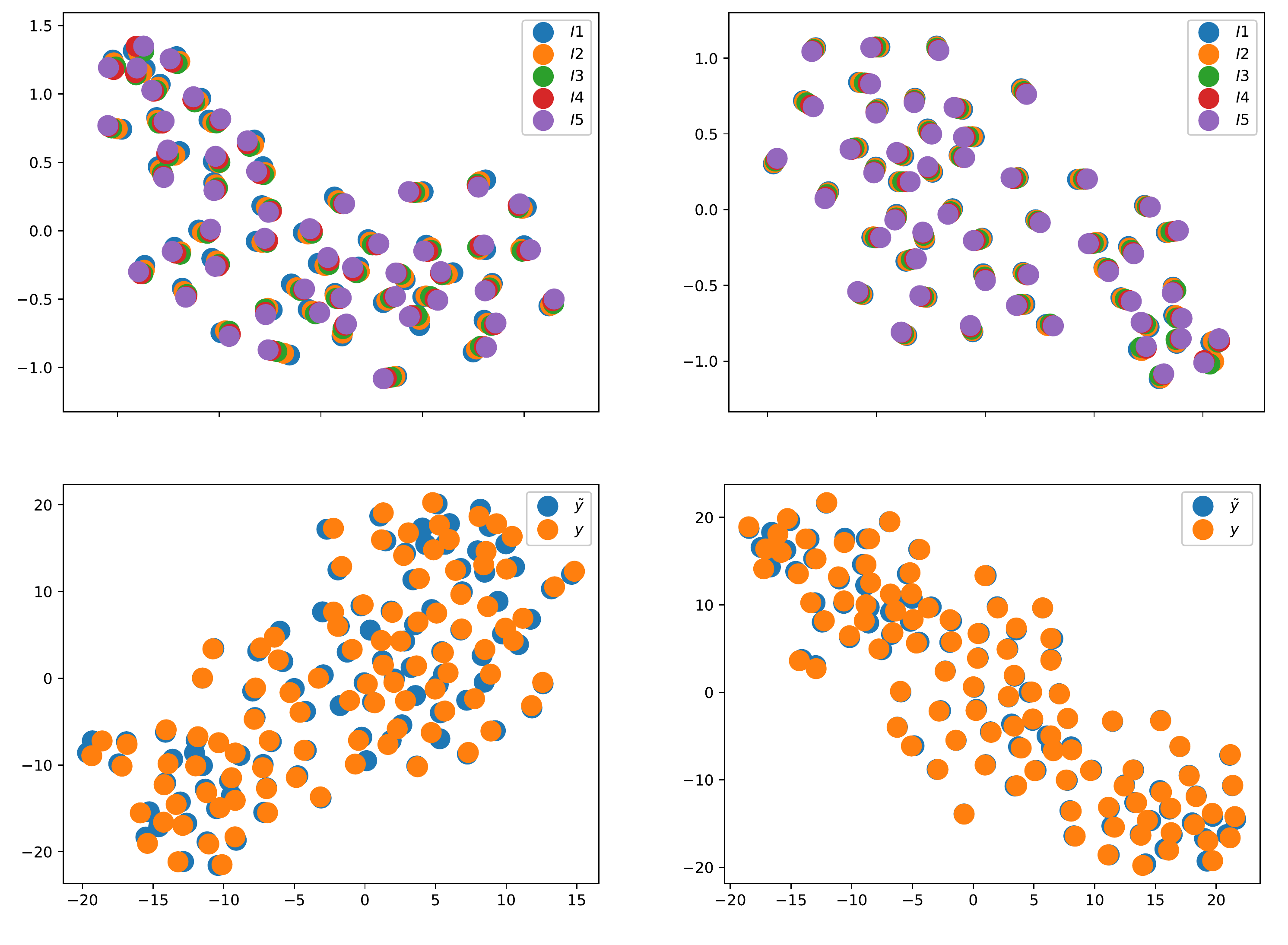}
\end{center}
\vspace{-0.5cm}
\caption{Top: a t-SNE based visualization of the latent space $r$  which represents the gaze angle (Sec.~\ref{GAM}).We show the corresponding latent spaces of GazeGAN~(top-left) and  GazeGANV2~(top-right). We plot 5 interpolated points ($I1-I5$) for each image and we use 50 images. Bottom: t-SNE visualization of the content features $c_{y}$ (orange) and $c_{\tilde y}$ (blue) extracted from $y$ and $\tilde y$, respectively. Bottom-Left: GazeGANV2 W/O $C$; Bottom-Right: GazeGANV2.}
\label{fig:AB_C}
\end{figure}

Fig.~\ref{fig:exp5} shows that our full-model (GazeGANV2) can better preserve identity information with respect to GazeGANV2 W/O $A$. To quantify this, we use $G_{x}$ to reconstruct test image samples $x$ ($x \in X$), and we measure the difference between the input image and the gaze-corrected result in local eye regions employing both MSSSIM and LPIPS metrics. Table~\ref{tab:exp2} shows that GazeGANV2 gets better scores than GazeGANV2 W/O $A$, confirming our design motivation. 

A shown in Fig.~\ref{fig:AB_A2}, we visualize the outputs of autoencoder with taking $y^{l}_{l}$, $y^{r}_{l}$, $F(y^{l}_{l})$ and $y^{r}_{l}$ as inputs after training. We can observe that the model can attain the similar reconstruction results for $y^{l}_{l}$ and $F(y^{r}_{l})$ as inputs, and can also attain the similar reconstruction results for $y^{r}_{l}$ and $F(y^{l}_{l})$ as inputs which validates the effectiveness of the objective loss.

{\bfseries Synthesis-As-Training Method.} The gaze animation results in Fig.~\ref{fig:exp2_1} and~\ref{fig:exp2_2} show the effectiveness of our method in disentangling the angle representation. 
Fig.~\ref{fig:AB_C} (top) shows a t-SNE visualization of points interpolated in the latent space. In more detail, 
following the procedure explained in Sec.~\ref{infer},
we uniformly interpolate the line connecting $r_{y}$ with $r_{y^{x}}$ in the angle latent space using 5 interpolation points ($I_1, ...I_5$) for each sample $y$.  Fig.~\ref{fig:AB_C} (top) shows that for each specific sample $y$, these five interpolation points are different from each other but strongly clustered together, which illustrates the disentanglement of the angle latent space.

\begin{table}[t]
\caption{Comparison between GazeGANV2 and GazeGANV2 W/O $D$ with respect to the generation quality in  gaze animation.}
\label{tab:AB_D_FID}
\centering
\begin{tabular}{lcccc}
\toprule
\multirow{2}{*}{Method} & \multicolumn{2}{c}{CelebGaze} & \multicolumn{2}{c}{CelebHQGaze} \\
\cmidrule(r){2-3}
\cmidrule(r){4-5} & GazeGANV2 & W/O $D$  & GazeGANV2 & W/O $D$ \\
\midrule
FID $\downarrow$ & {\bfseries 53.32} & 78.32 & {\bfseries 71.37} & 74.04 \\
\bottomrule
\end{tabular}
\end{table}

\begin{table}[t] \small
\centering
\caption{Quantitative comparison between CFM of GazeGANV2 with Bilinear and super-resolution model ESRGAN~\cite{wang2018esrgan}.
}
\begin{tabular}{ccccc}
\toprule
Metrics & Bilinear & ESRGAN & CFM \\
\midrule
MSSSIM $\uparrow$  & 0.9563 & 0.9595  & {\bfseries 0.9827} \\
LPIPS $\downarrow$  & 0.2393 & 0.1476 & {\bfseries 0.1039} \\  
FPS $\downarrow$  & 30.600 & 4.3000 &  {\bfseries 27.700} \\ 
Params $\downarrow$  & {\bfseries 48.88M} & 80.88M & 49.23M \\ 
\bottomrule  
\end{tabular}
\label{tab:AB_D_ALL}
\vspace{-0.4cm}
\end{table}

{\bfseries Latent Reconstruction Loss $\mathcal{L}_{fp}$.} We use 
$G_{y}$ to fill in the eye region of test images $y$ ($y \in Y$), and we measure the difference between the input images and the generated results employing MSSSIM and LPIPS. In Table~\ref{tab:AB_C}, GazeGANV2 obtains better scores than GazeGANV2 W/O $C$, which shows that $\mathcal{L}_{fp}$ further improves the ability to preserve identity information. Moreover, we visualize the content features $c_{y}$ and $c_{\tilde y}$ extracted from real samples $y$ and reconstructed samples $\tilde y$ across all the $Y$ test data. As shown in Fig.~\ref{fig:AB_C} (bottom), we observe that using our full model GazeGANV2, $c_{y}$ and $c_{\tilde y}$ usually lie closer to each other to what happens when using GazeGANV2 W/O $C$, and it shows that this loss helps to represent content information consistently.

{\bfseries Coarse-to-Fine Module (CFM).} The previous experiments validate the effectiveness of CFM. As shown in the 2nd and the 4th row of Fig.~\ref{fig:exp5}, the gaze correction results of GazeGANV2 
are more realistic than those produced by GazeGANV2 W/O $D$. In Table~\ref{tab:AB_D_FID}, the quantitative comparison between GazeGAN and GazeGAN W/O $D$ confirms the effectiveness of CFM. Then, Fig.~\ref{fig:AB_D2} shows the differences in the gaze correction results obtained with GazeGAN and GazeGAN W/O $D$.
The heatmap of the difference between the two generated images shows that CFM can compensate for the high-frequency information loss of the coarse output. Finally, we compare our CFM with some upsampling methods, such as Bilinear and super-resolution method, ESRGAN~\cite{wang2018esrgan}. By taking all samples $y^{h}$ from domain $Y^{h}$, we attain all low-resolution reconstructed results $\tilde y$. Then, three different methods are used for upsampling them to attain high-resolution results. Fig.~\ref{fig:AB_D3} shows our method achieves better reconstruction with fewer artifacts, such as eye regions. Quantitative experiments of Table~\ref{tab:AB_D_ALL} show our CFM achieves better MSSSIM and LPIPS scores and has higher FPS and fewer parameters than ESRGAN.

\section{Conclusion}
In this paper, we introduce a new high-resolution gaze dataset in the wild, CelebHQGaze, which is characterized by a large diversity in head poses and gaze angles. Moreover, we propose a novel unsupervised method, GazeGANV2, for gaze-direction correction and animation. GazeGANV2 formulates the gaze correction problem as an inpainting task and uses a coarse-to-fine learning strategy to generate high-resolution images. Moreover, self-supervised learning and Synthesis-As-Training methods are used to disentangle the content and angle-specific features, which can condition the generation process. The qualitative and quantitative results demonstrate the method's effectiveness and its superiority to the state of the arts.

\ifCLASSOPTIONcaptionsoff
  \newpage
\fi

\bibliographystyle{ieeets}
\bibliography{gazegan}

\end{document}